\documentclass[10pt,twocolumn,letterpaper]{article}

\usepackage[pagenumbers]{cvpr} 

\usepackage{amsmath,amsfonts,bm}
\usepackage{xspace}

\providecommand{\eg}{\textit{e.g.}\@\xspace}
\providecommand{\ie}{\textit{i.e.}\@\xspace}

\def\eqref#1{equation~\ref{#1}}

\def\1{\bm{1}}

\def\rmM{{\mathbf{M}}}

\def\rmX{{\mathbf{X}}}

\def\rmZ{{\mathbf{Z}}}

\def\vz{{\bm{z}}}

\DeclareMathAlphabet{\mathsfit}{\encodingdefault}{\sfdefault}{m}{sl}
\SetMathAlphabet{\mathsfit}{bold}{\encodingdefault}{\sfdefault}{bx}{n}

\def\sR{{\mathbb{R}}}

\usepackage{multirow}

\usepackage{xcolor}         
\usepackage{amsmath}
\usepackage{bm}
\usepackage{graphicx}
\usepackage{svg}
\usepackage{multirow}
\usepackage{colortbl}
\usepackage{wrapfig}
\usepackage{pifont}
\usepackage{subcaption}
\newcommand{\cmark}{\ding{51}}%
\newcommand{\xmark}{\ding{55}}%
\definecolor{ourscolor}{HTML}{c2d1e5}

\definecolor{cvprblue}{rgb}{0.21,0.49,0.74}
\usepackage[pagebackref,breaklinks,colorlinks,allcolors=cvprblue]{hyperref}

\def\paperID{482} 
\def\confName{CVPR}
\def\confYear{2025}

\title{F-LMM: Grounding Frozen Large Multimodal Models}

\author{
\centerline{
Size Wu\textsuperscript{\rm 1}\quad
Sheng Jin\textsuperscript{\rm 2}\quad
Wenwei Zhang\textsuperscript{\rm 3}\quad
Lumin Xu\textsuperscript{\rm 4}\quad
Wentao Liu\textsuperscript{\rm 2,3}\quad
Wei Li\textsuperscript{\rm 1}\quad
Chen Change Loy\textsuperscript{\rm 1}
} \\
\centerline{
\textsuperscript{\rm 1} S-Lab, Nanyang Technological University \quad
\textsuperscript{\rm 2} SenseTime Research and Tetras.AI
}\\
\centerline{
\textsuperscript{\rm 3} Shanghai AI Laboratory
 \quad 
\textsuperscript{\rm 4} The Chinese University of Hong Kong 
}\\
\centerline{
\url{size001@e.ntu.edu.sg} \qquad \url{{wei.l,ccloy}@ntu.edu.sg}
}
}

\begin{document}


\twocolumn[{%
\renewcommand\twocolumn[1][]{#1}%
\maketitle
\vspace{-8mm}
\centering
\includegraphics[width=1.0\textwidth]{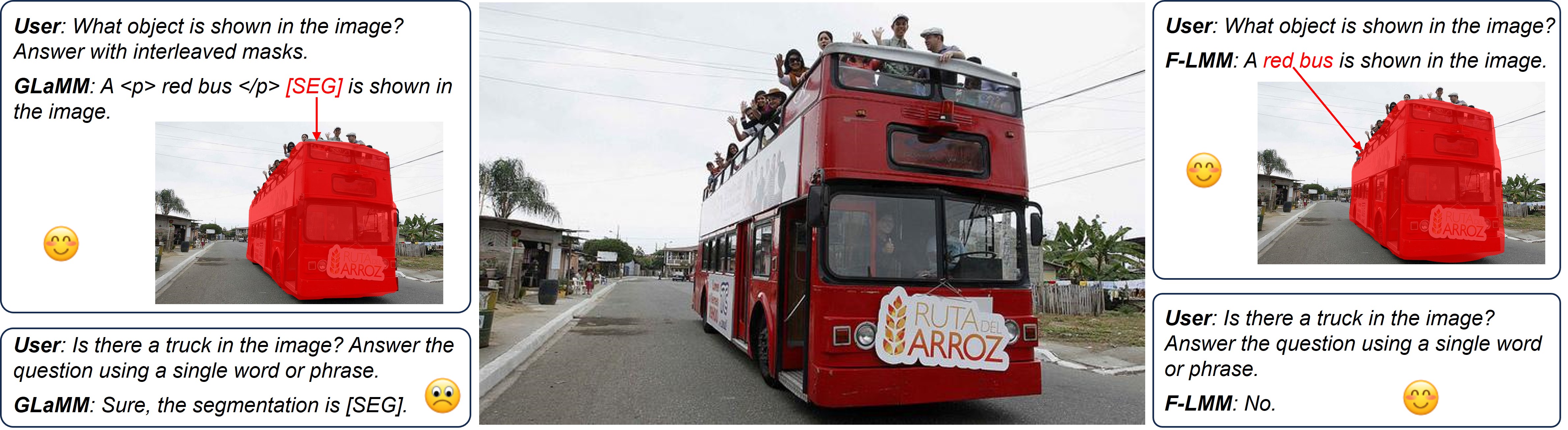}
\vspace{-3mm}
\captionof{figure}{An example of user-AI conversation around an image. \textbf{Left}: The current state-of-the-art grounding model GLaMM~\cite{hanoona2023GLaMM} is effective for grounded conversation when prompted by \textit{"answer with interleaved masks"}, but fails to follow user instruction to answer a single word (yes or no) and misunderstands the question as a referring segmentation prompt. \textbf{Right}: Our F-LMM preserves instruction-following ability while being able to perform visual grounding.}
\label{fig:flmm_teaser1}
\vspace{0.7cm}
}]

\begin{abstract}
Endowing Large Multimodal Models (LMMs) with visual grounding capability can significantly enhance AIs' understanding of the visual world and their interaction with humans. However, existing methods typically fine-tune the parameters of LMMs to learn additional segmentation tokens and overfit grounding and segmentation datasets. Such a design would inevitably cause a catastrophic diminution in the indispensable conversational capability of general AI assistants. In this paper, we comprehensively evaluate state-of-the-art grounding LMMs across a suite of multimodal question-answering benchmarks, observing drastic performance drops that indicate vanishing general knowledge comprehension and weakened instruction following ability. To address this issue, we present F-LMM---grounding \emph{frozen} off-the-shelf LMMs in human-AI conversations---a straightforward yet effective design based on the fact that word-pixel correspondences conducive to visual grounding inherently exist in the attention mechanism of well-trained LMMs. Using only a few trainable CNN layers, we can translate word-pixel attention weights to mask logits, which a SAM-based mask refiner can further optimise. Our F-LMM neither learns special segmentation tokens nor utilises high-quality grounded instruction-tuning data, but achieves competitive performance on referring expression segmentation and panoptic narrative grounding benchmarks while completely preserving LMMs' original conversational ability. Additionally, with instruction-following ability preserved and grounding ability obtained, F-LMM can be directly applied to complex tasks like reasoning segmentation, grounded conversation generation and visual chain-of-thought reasoning. Our code can be found at \url{https://github.com/wusize/F-LMM}. 
\end{abstract}    
\section{Introduction}
Large Multimodal Models (LMMs), which integrate Large Language Models (LLMs) with visual signals, have demonstrated remarkable success in multimodal understanding, reasoning and interaction~\cite{liu2023llava, liu2023improvedllava, liu2024llavanext, lu2024deepseekvl, li2024mgm, hpt,zang2023contextual}. To further advance LMMs with better perception capability, a recent line of research~\cite{zhang2023llavagrounding, lai2023lisa, hanoona2023GLaMM, ren2023pixellm, wei2024lasagna, zhang2024groundhog} that visually grounds language contents in user-model conversations has drawn increasing attention. This explicit association between key phrases/words and visual objects greatly enhances LMMs' understanding of the visual world and allows for more intuitive and meaningful human-AI interactions.

\begin{figure*}[t]
  \centering
\vspace{-5mm}
\includegraphics[width=1.0\textwidth]
{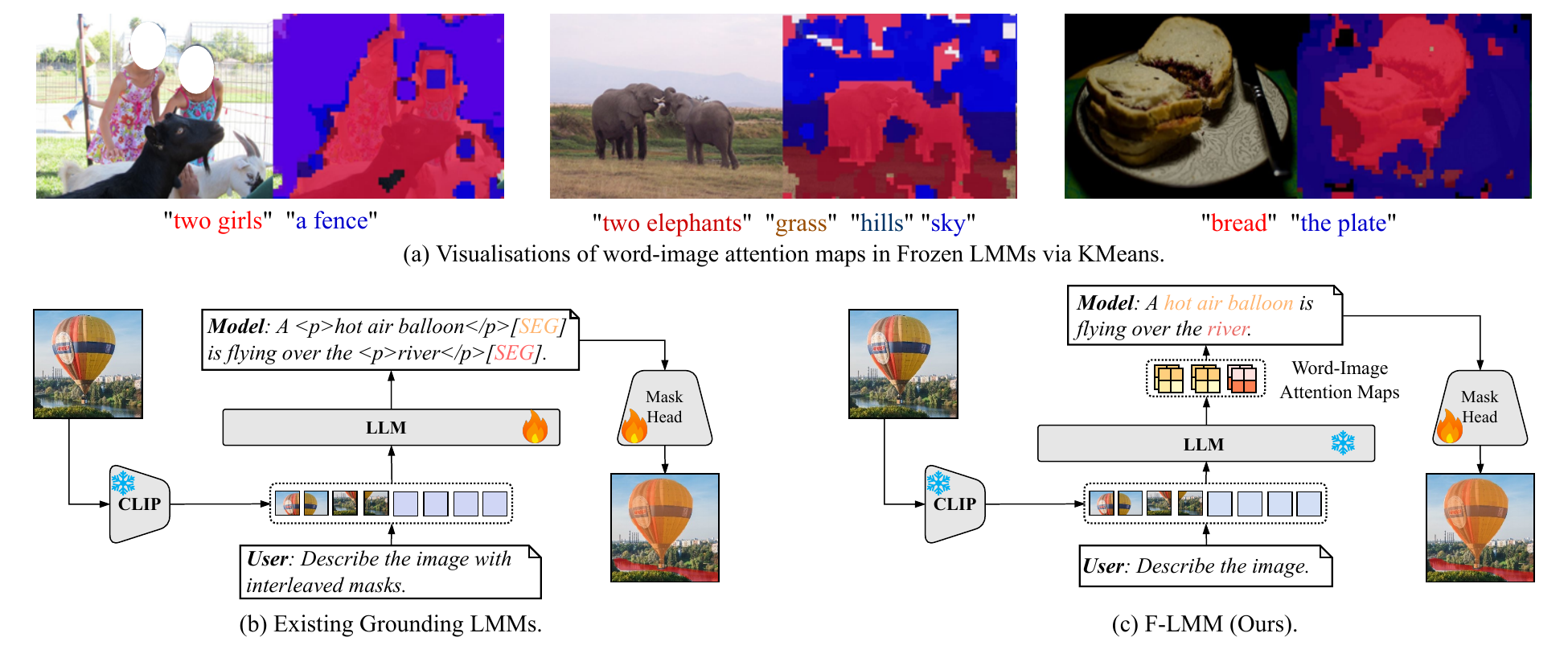}
\vspace{-16pt}
  \caption{(a) Geometric and spatial cues conducive to visual grounding are observed in the visualisations of word-image attention maps in frozen LMMs. (b) Existing grounding LMMs are fine-tuned to generate a special mask token (\eg, \texttt{[SEG]}) for visual grounding purposes, which ruins the original conversational ability. (c) Our F-LMM translates word-image attention maps from frozen LMMs to grounding masks, while fully preserving the general-purpose chat capability.
    }
  \label{fig:flmm_teaser23_merge}
  \vspace{-6pt}
\end{figure*}

By design, one commonly adopted build (Figure~\ref{fig:flmm_teaser23_merge}(b)) for visually grounding language contents is connecting LMMs with a mask head (\eg, Segment Anything Model (SAM)~\cite{kirillov2023sam}), wherein both the LLM backbone and the mask head are fine-tuned with well-prepared visual grounding data that contains segmentation annotations. Also, some additional learnable tokens (\eg, \texttt{[SEG]}) are introduced to the LMMs' vocabulary, to directly associate key phrases or words with visual objects in conversations. However, this design will inevitably provoke a \emph{catastrophic diminution} in general knowledge comprehension and instruction-following ability due to the following
reasons. First, existing segmentation and visual grounding data only contain \emph{elementary} patterns for answering simple grounding prompts. Second, during the fine-tuning stage, the LMMs are mainly optimised for effectively modelling the relationship between key phrases or words and special segmentation tokens, \ie, overfitting the segmentation and grounding data. Therefore, the conversational ability is sacrificed. For instance, the state-of-the-art grounding model GLaMM~\cite{hanoona2023GLaMM} fails to answer a simple yes-or-no question (Figure~\ref{fig:flmm_teaser1}). Moreover, quantitative evaluations of existing grounding LMMs in conversational ability are presented in Table~\ref{tab:benchmarks}, with zero or near-zero scores on general multimodal question-answering benchmarks necessitating instruction-following ability.

One possible option to deal with this dilemma is to collect high-quality training data encompassing both meaningful conversations and mask annotations. For example, LLaVA-G~\cite{zhang2023llavagrounding} annotates the 150k LLaVA-Instruct data samples~\cite{liu2023llava} with segmentation masks so that the LMMs simultaneously learn to chat and segment. Nonetheless, annotating high-quality grounded conversation data is costly and hard to scale. Despite being trained on costly annotated data, LLaVA-G still lags behind general-purpose LMMs on multimodal understanding tasks. Furthermore, training on large-scale annotated data normally consumes significant computational resources, which is, obviously not a resource-efficient solution.

In this paper, we propose a simple yet effective design, \ie, grounding frozen LMMs (dubbed as F-LMM) in human-AI conversations. We argue that freezing the parameters of well-trained LMMs is \emph{the most practical} design choice for fully preserving the original excellent conversational ability when building general-purpose grounding LMMs. In particular, we take inspiration from the built-in interpretability of the attention mechanism in transformers~\cite{transformer,chefer2021generic} that represents interrelations between text and image contents in design. We observe that off-the-shelf LMMs already produce word-pixel correspondences necessary for visual grounding, despite they were not explicitly pre-trained with region or pixel annotations. As illustrated in Figure~\ref{fig:flmm_teaser23_merge}(a), we visualise word-image attention maps from frozen LMMs via K-Means clustering, revealing notable geometric and spatial cues of the objects~\footnote{For better visibility, we perform K-Means clustering on the stack-up of all attention maps collected in a forward pass instead of selecting a single attention map.}. For example, coarse visual grounding masks for key phrases (\eg, ``two girls", ``two elephants", and ``the plate") in language sentences emerge from attention maps in LMMs. Therefore, our F-LMM takes these visual-language correspondences as useful segmentation priors for decoding grounding masks, without further tuning the LMMs' weights or learning a special segmentation token to model object locations, as shown in Figure~\ref{fig:flmm_teaser23_merge}(c).

The only trainable part of our F-LMM is a mask head plus a keyword selector. The mask head comprises a CNN-based mask decoder (a tiny U-Net~\cite{ronneberger2015u}) that translates stacked attention maps to mask logits and a light-weight mask refiner (retrofitted from  SAM~\cite{kirillov2023sam}'s mask head) that uses additional image and language cues to refine the semantic-agnostic masks from the mask decoder. The keyword selector is a linear layer that discovers object nouns in text sequences, automating the process of grounded human-AI conversation.
Moreover, we only use the RefCOCO(+/g)~\cite{refcoco, refcocog} and PNG~\cite{png} datasets as our training data, enabling LMMs to segment user-described objects and ground key phrases or words in a text sequence. Unlike previous works~\cite{zhang2023llavagrounding,hanoona2023GLaMM,ren2023pixellm}, our F-LMM eliminates the necessity for high-quality conversation data that are annotated with masks to preserve conversational ability when learning grounding.

Our experiments demonstrate that F-LMM maintains the original excellence of off-the-shelf LMMs on general question-answering benchmarks, while achieving competitive results on referring segmentation and phrase grounding. In more complex tasks like reasoning segmentation, grounded conversation generation and visual chain-of-thought reasoning, F-LMM achieves better or comparable results when contrasted with models specially trained for such tasks. Compared with existing grounding LMMs, F-LMM offers the best balance between grounding and chat capabilities.

\section{Related Work}
\noindent\textbf{Large Multimodal Models.}
Recent advancements in LMMs
~\cite{alayrac2022flamingo, li2023blip2, dai2024instructblip, liu2023llava, liu2023improvedllava, liu2024llavanext, ye2023mplug, bai2023qwenvl, li2023otter, lin2023vila, mckinzie2024mm1, lu2024deepseekvl, li2024mgm, hpt, laurenccon2024matters} have been fueled by the success of LLMs~\cite{brown2020gpt3, achiam2023gpt4, zhang2022opt, llama, llama2, jiang2023mistral, chowdhery2023palm, vicuna, mesnard2024gemma} since the debut of GPT series~\cite{radford2018gpt, radford2019gpt2, brown2020gpt3, achiam2023gpt4} that feature an auto-regressive framework based on transformer decoders~\cite{transformer}. These LLMs possess general world knowledge and excellent conversational ability to follow human instructions, thanks to large-scale generative pre-training~\cite{brown2020gpt3} and supervised finetuning on instruction-tuning data~\cite{wei2021finetuned} or human feedback~\cite{ouyang2022hf}. 
By integrating image representations from vision encoders~\cite{radford2021learning, zhai2023sigmoid} to LLMs, LMMs enable visual understanding and reasoning in AI assistants. This integration is usually established by a multilayer perceptron (MLP) that directly maps image features to the LLMs' input embedding space~\cite{liu2023llava, liu2023improvedllava, liu2024llavanext, lu2024deepseekvl, li2024mgm, hpt,zang2023contextual} or a cross-attention module that abstracts the image contents with a set of query embeddings~\cite{alayrac2022flamingo,li2023blip2,bai2023qwenvl,ye2023mplug}. In our research, we build F-LMM on LMMs of the former type (MLP-based), which preserves images' 2-D topological structure in the cross-modal integration.

\noindent\textbf{Visual Segmentation.}
The task of predicting 2D masks for visual objects is known as image segmentation, which can be categorised into semantic segmentation~\cite{chen2017deeplab, cocostuff, zhou2017ade20k, cheng2021maskformer}, instance segmentation~\cite{he2017mask, cheng2022mask2former, zhang2021knet} and panoptic segmentation~\cite{panoptic, cheng2020panoptic_deeplab, xiong2019upsnet_panopticseg, li2021fcn4panoptic_seg} depending on whether the goal is to differentiate pixel semantics or object instances. 
These standard segmentation approaches rely on a pre-defined set of object classes for recognition. In contrast, referring expression segmentation (RES)~\cite{refcoco,refcocog,nagaraja2016modeling_ref,zou2023generalized, luo2020mcn, yang2022lavt, liu2023gres} involves segmenting objects based on free-form human language descriptions, allowing for enhanced human-model interaction. Additionally, panoptic narrative grounding (PNG)~\cite{png, ding2022ppmn, wang2023epng,  guo2024xpng} requires segmenting masks for key phrases or words in a sentence. In this study, we mainly leverage RES and PNG tasks to evaluate the grounding capability of LMMs. Besides, we also test LMMs' segmentation ability in complex scenarios that necessitate reasoning~\cite{hanoona2023GLaMM, lai2023lisa, shao2024visualcot}. Moreover, the prompt-based SAM~\cite{kirillov2023sam} pre-trained on billion-scale high-quality mask data has become a constituent component in many grounding LMMs to boost segmentation performance. We also adopt SAM's mask head to initialise our mask refiner.

\noindent\textbf{Grounding Large Multimodal Models.}
Grounding Large Multimodal Models~\cite{peng2023kosmos2,chen2023shikra,bai2023qwenvl,you2023ferret,zhang2023llavagrounding,zhang2024groundhog,ren2023pixellm,hanoona2023GLaMM,wei2024lasagna,lai2023lisa,xia2024gsva,pi2023perceptiongpt,zang2023contextual,zhang2024psalm} can localise language contents during user-model conversations. 
Some approaches~\cite{peng2023kosmos2, chen2023shikra, you2023ferret, bai2023qwenvl} represent coordinates of bounding boxes as texts and train LMMs to predict the coordinates in a generative manner. Several recent works~\cite{lai2023lisa, zhang2023llavagrounding, ren2023pixellm, wei2024lasagna, hanoona2023GLaMM, pi2023perceptiongpt} train LMMs to predict a special segmentation token for encoding the grounded object and utilise a segmentation head (\eg, SAM~\cite{kirillov2023sam}) to decode object masks. 
This study mainly focuses on grounding LMMs with segmentation ability for visual perception. 
To obtain competitive visual grounding performance, existing works extensively fine-tune the parameters of LMMs on a large amount of segmentation~\cite{zhou2017ade20k, cocostuff, panoptic, ramanathan2023paco, he2022partimagenet} and grounding~\cite{refcoco, refcocog, kazemzadeh2014referitgame, visualgenome, plummer2015flickr30k, png} datasets. And to balance the LMMs' grounding and conversational abilities, there are efforts~\cite{zhang2023llavagrounding, ren2023pixellm, zhang2024groundhog} to collect high-quality instruction-tuning data annotated with segmentation masks. In contrast, we make the first attempt to build grounding LMMs on top of off-the-shelf LMMs without fine-tuning their parameters. Furthermore, we bypass the need for grounded instruction-tuning data to preserve decent chat ability. 

\section{Method}

\begin{figure*}[ht]
  \centering
\includegraphics[width=1.0\textwidth]{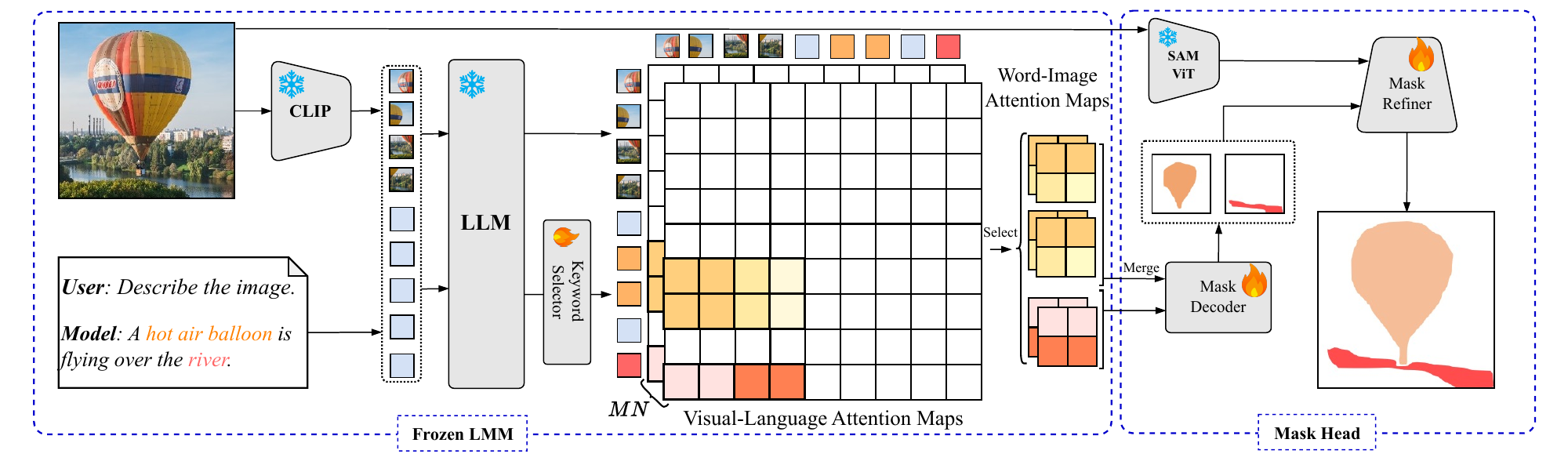}
\vspace{-16pt}
  \caption{
   The overall pipeline of F-LMM. The word-image attention maps from the frozen LLM serve as segmentation priors for the mask head. The keyword selector discovers object nouns in the text sequence. The mask head encompasses a mask decoder that translates attention weights to mask logits and a mask refiner that optimises the mask decoder's predictions. $M$ and $N$  represent the numbers of transformer layers and attention heads. 
  }
  \vspace{-4pt}
  \label{fig:method}
\end{figure*}

\begin{figure}[t]
  \centering
\includegraphics[width=0.45\textwidth]{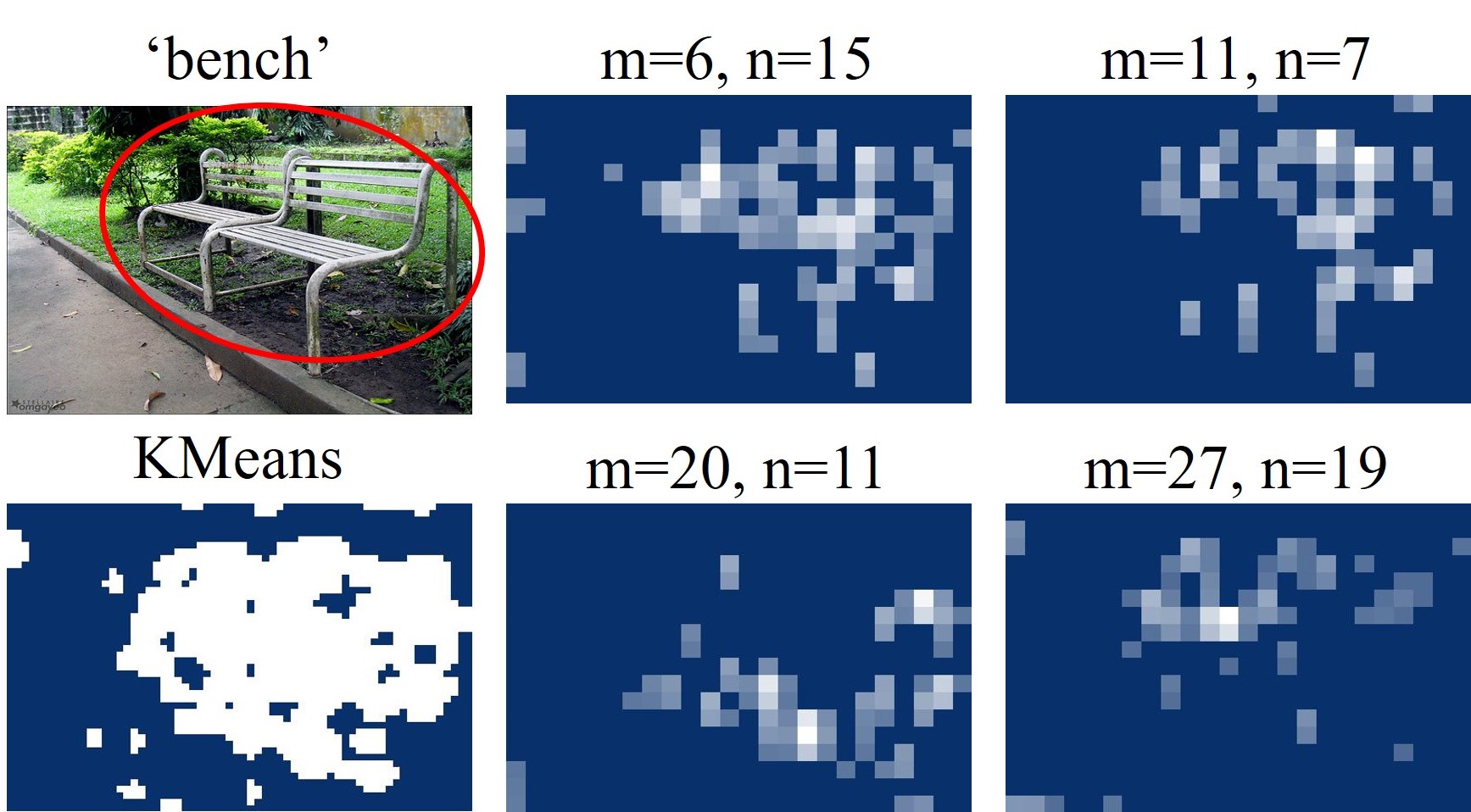}
  \caption{Visualisations of word-image attention maps. The letters $m$ and $n$ indicate that the attention map is derived from the $n$-th attention head of the $m$-th transformer layer. 
  }
  \label{fig:attention_heads_layers}
\end{figure}

In this section, we introduce our F-LMM by first probing the causal attention mechanism in LMMs with visualisations of word-image attention maps in Sec~\ref{sec:segmentation_priors}. Then, we elaborate on F-LMM exploiting segmentation priors from frozen LMMs for visual grounding using the mask head in Sec~\ref{sec:mask_head}. Finally, we show how to automate the process of grounded conversation with a linear keyword selector to indicate words of grounding targets in Sec~\ref{sec:grounding_target_selection}. The overall pipeline is illustrated in Figure~\ref{fig:method}.


\subsection{Segmentation Priors from Frozen LMM}\label{sec:segmentation_priors}

\noindent\textbf{Vision-Language Sequence.} A typical build of a LMM~\footnote{In this paper, the term `multimodal' stands for vision and language modalities.} comprises an image encoder $f_{v}$ (\eg, CLIP~\cite{radford2021learning}~\footnote{The image encoder might be any vision model that is pre-trained on image-text pairs. We use the classic term `CLIP' in this paper to represent all such models for brevity.}), a vision-language projector $f_p$, and a LLM, denoted as $f_{\mathrm{llm}}$. The inputs to an LMM are usually an image $\rmX_{v} \in \sR^{3 \times H \times W}$ and the associated text $\rmX_{t}$. The input image is first encoded by the vision encoder $f_{v}$ and then mapped to the input space of the LLM by the projector $f_p$:
\begin{equation*}
\rmZ_v = f_p(\texttt{Flatten}(f_{v}(\rmX_{v}))) \in \sR^{hw \times d},
\end{equation*}
where $h$ and $w$ are the height and width of projected feature maps via $f_{v}$. The $\texttt{Flatten}$ operation unfolds the 2-D image feature map to a 1-D sequence. The constant $d$ is the hidden state dimension of the LLM.
Likewise, the text input is first encoded as discrete tokens and then mapped to text embeddings:
\begin{equation*}
     \rmZ_t = \texttt{Embed}(\texttt{Tokenize}(\rmX_{t})) \in \sR^{L \times d},
\end{equation*}
where $L$ denotes the length of text embeddings. The visual-language sequence input to the LLM is a concatenation of image and text embeddings: $\rmZ = \{\rmZ_v, \rmZ_t\} \in \sR^{(hw+L) \times d}$.

\begin{table*}[t]
  \centering
\caption{The main evaluation results on question-answering benchmarks, referring expression segmentation (RES) benchmark and panoptic narrative grounding (PNG) benchmark. LLaVA$^{\mathrm{W}}$: LLaVA-In-the-Wild. LLaVA-1.6 and MGM-HD take high-resolution image inputs. LLaVA-1.6-M-7B means the model is based on Mistral-7B~\cite{jiang2023mistral}. GLaMM-FS-7B means we use the `FullScope' version of GLaMM.}
\scalebox{0.8}
{\begin{tabular}{l|cccc|ccc|ccc}
  \toprule
 \multirow{2}{*}{Model}&\multicolumn{4}{c|}{Multimodal Question Answering} &\multicolumn{3}{c|}{RES}&\multicolumn{3}{c}{PNG}  \\
&MME&MMBench&MMVet&LLaVA$^{\mathrm{W}}$&RefCOCO&RefCOCO+& RefCOCOg&All & Thing & Stuff\\ \hline

\hline
PixelLM-7B~\cite{ren2023pixellm}&309/135&17.4&15.9&46.4&
73.0&66.3&69.3&43.1&41.0&47.9\\
LISA-7B~\cite{lai2023lisa}&1/1&0.4&19.1&47.5&74.9&65.1&67.9&-&-&-\\
PerceptionGPT-7B~\cite{pi2023perceptiongpt}&-&-&-&-& 75.1 & 68.5 & 70.3&-&-&- \\
LLaVA-G-7B~\cite{zhang2023llavagrounding}&-&-&-&55.8&77.1&68.8&71.5&-&-&-\\
GroundHog-7B~\cite{zhang2024groundhog}&-&-&-&-&78.5&70.5&74.1&66.8&65.0&69.4\\
GLaMM-FS-7B~\cite{hanoona2023GLaMM}&14/9&36.8&10.3&32.0&
78.6&70.5&74.8&55.8&52.9&62.3\\

GSVA-7B~\cite{xia2024gsva}&446/18&17.8&19.4&38.3 &77.7 & 68.2 & 73.2 & 41.8&39.6&46.6\\

LaSagnA-7B~\cite{wei2024lasagna}&0/0&0.0&16.7&34.5&76.8&66.4&70.6&-&-&-\\ \hline
F-LMM (DeepSeekVL-1.3B~\cite{lu2024deepseekvl})&1307/225&64.6&34.8&51.1&75.0&62.8&68.2&64.9&63.4&68.3
\\
F-LMM (MGM-2B~\cite{li2024mgm})&1341/312&59.8 &31.1&65.9&75.0& 63.7& 67.3&65.6&64.4&68.4\\ 
F-LMM (LLaVA-1.5-7B~\cite{liu2023improvedllava})&1511/348&64.3&30.5&69.0&75.2&63.7&67.1&64.8&63.4&68.2\\
F-LMM (HPT-Air-6B~\cite{hpt})&1010/
258&69.8&31.3&59.2&74.3&64.0&67.5&65.5&64.0&68.8\\ 
F-LMM (HPT-Air-1.5-8B~\cite{hpt})&1476/308&75.2&36.3&62.1&76.3&64.5&68.5&65.4&64.1&68.5\\ 
F-LMM (MGM-7B~\cite{li2024mgm})&1523/316&69.3&40.8&75.8&75.7&64.8&68.3&66.3&65.3&68.6\\
F-LMM (DeepSeekVL-7B~\cite{lu2024deepseekvl})&1468/298&73.2&41.5&77.8&76.1&66.4&70.1&65.7&64.5&68.5
\\   
F-LMM (LLaVA-1.6-7B~\cite{liu2024llavanext})&1519/322&68.1&44.1&72.3&75.8&65.8&70.1&66.3&65.1&69.0\\
F-LMM (LLaVA-1.6-M-7B~\cite{liu2024llavanext})&1501/324&69.5&47.8&71.7&75.7&66.5&70.1&66.5&65.4&69.1\\
F-LMM (MGM-HD-7B~\cite{li2024mgm})&1546/319&65.8&41.3&74.0&76.1&65.2&68.5&66.7&65.6&69.1\\
\bottomrule
\end{tabular}
}
\label{tab:benchmarks}
\end{table*}

\noindent\textbf{Segmentation Priors in Self-Attention.}
The vision-language sequence is mainly processed by causal self-attentions~\cite{transformer, radford2018gpt} in the LLM, including inner product and weighted-sum operations. Specifically, for a word token with position index $i$ in the vision-language sequence $\rmZ$ , its embedding $\vz^i$ is updated by the weighted sum of the first $i$ embeddings: $\hat{\vz}^{i} = \texttt{SoftMax}(\frac{\vz^i \cdot \rmZ[:i]}{d}) \cdot \rmZ[:i]$, where $\texttt{SoftMax}(\frac{\vz^i \cdot \rmZ[:i]}{d})$ is the attention weights. Here, we omit the residual layers and feedforward layers for brevity.

Considering the word-image interaction in the multimodal scenario, we can select the word token's attention weights with the image embeddings from the overall vision-language attention weights:
\begin{equation*}\label{eq:attention_map}
    \bm{a}^i = \texttt{Unflatten}(\texttt{SoftMax}(\frac{\vz^i \cdot \rmZ[:i]}{d})[:hw]) \in \sR^{h\times w},
\end{equation*}
where $\texttt{Unflatten}$ restores the 2-D spatial structure from the 1-D sequence to form an attention map. In Figure~\ref{fig:attention_heads_layers}, we visualise such word-image attention maps from various transformers layers and attention heads in an LMM (\ie, DeepSeekVL-7B~\cite{lu2024deepseekvl}). The objects' shape and location can be observed in word-image attention maps of certain layers or heads. The visibility is further enhanced when we stack the attention maps from all layers and heads and perform K-Means clustering.
It can be observed that the attention maps offer meaningful \textit{segmentation priors} with spatial and geometric cues for grounding objects visually.

\noindent\textbf{Language Cues.} In addition to the spatial and geometric cues from word-image attention maps, F-LMM can also capitalise on the object's corresponding text embeddings from the LLM $f_{\mathrm{llm}}$, which provide extra language cues for the grounding of visual objects.

\subsection{Visual Grounding with Mask Head}\label{sec:mask_head}
We use the segmentation priors from the frozen LMM for pixel-level grounding, with the help of a mask head consisting of a mask decoder and a mask refiner.

\noindent\textbf{Mask Decoder.} 
The mask decoder $f_d$ is a 2-D CNN model that transforms the word-image attention maps of grounded objects into mask logits, which is instantiated by a 3-stage U-Net~\cite{ronneberger2015u}. Please refer to the supplemental material for details on the mask decoder. The extraction of word-image attention map $\bm{a}^i$ for a word token with position index $i$ is illustrated in Eq.~\ref{eq:attention_map} and Figure~\ref{fig:method}.
For an object described by multiple words, we merge their corresponding word-image attention maps to a single attention map $\bm{a}$ via element-wise average or max operation. The attention map $\bm{a}$ is further normalised as $\bm{a} / \mathrm{sum}(\bm{a})$ so that all elements sum to 1. Considering $M$ layers and $N$ attention heads, we stack the $MN$ attention maps as $\bm{A} \in \sR^{MN\times h\times w}$, which forms the input to a mask decoder. Given the importance of high input resolution for segmentation models, we upsample the stacked attention maps $\bm{A}$ to $h' \times w'$ by bilinear interpolation before feeding it to a mask decoder, where $h' > h$ and $w' > w$. In practice, we set $h'=w'=64$. Then, the mask decoder maps $\bm{A}$ into mask logits: $\rmM_{\mathrm{logits}} = f_d(\bm{A})$. 
We derive the corresponding binary mask via $\rmM_{\mathrm{pred}}=\rmM_{\mathrm{logits}} > 0$. During training, the mask decoder is optimised with BCE and DICE losses~\cite{sudre2017dice}.

\begin{table*}[t]
\hspace{0.033\textwidth}
\begin{minipage}[t]{0.33\textwidth}
\centering
\caption{Reasoning Segmentation.}
\vspace{-6pt}
\scalebox{0.8}{\begin{tabular}{l|c|ccc}
  \toprule
 \multirow{2}{*}{Model}&Val&\multicolumn{3}{c}{Test}  \\
&-&Short & Long & All \\ \hline
X-Decoder~\cite{zou2023generalized}&22.6&20.4&22.2&21.7\\
SEEM~\cite{zou2024seem}&25.5&20.1&25.6&24.3\\
GroundingSAM~\cite{liu2023grounding}&26.0&17.8&22.4&21.3\\
OVSeg~\cite{liang2023open}&28.5&18.0&28.7&26.1\\
LISA~\cite{lai2023lisa}&44.4&{37.6}&36.6&36.8\\
F-LMM&{46.7}&36.9&{49.1}&{46.2}\\
\bottomrule
\end{tabular}
}
\label{tab:reason_seg}
\end{minipage}
\begin{minipage}[t]{0.65\textwidth}
\centering
\caption{Grounded Conversation Generation (GCG). M. stands for METEOR. }
\vspace{-6pt}
\scalebox{0.8}
{\begin{tabular}{l|c|ccc|ccc}
  \toprule
 \multirow{2}{*}{Model}&GCG&\multicolumn{3}{c|}{Val}&\multicolumn{3}{c}{Test}  \\
&Training&M.&mIoU& Recall&M. & mIoU & Recall\\ \hline
LISA~\cite{lai2023lisa}&\cmark&13.0 &62.0 &36.3 &12.9 &61.7 &35.5\\
OMG-LLaVA~\cite{zhang2024omg}&\cmark&14.9&65.5&-&14.5&64.7&-\\
GLaMM~\cite{hanoona2023GLaMM}&\cmark&16.2 
&{66.3} &41.8 &15.8 &{65.6}&{40.8}\\
\hline
BuboGPT~\cite{zhao2023bubogpt}&\xmark&17.2&54.0 &29.4 &17.1 &54.1&27.0\\
KOSMOS-2~\cite{peng2023kosmos}&\xmark&16.1& 55.6&28.3 &15.8 &56.8 &29.0\\
F-LMM&\xmark&{17.6}&{63.5}&{42.0}&{17.4}&{63.6}&{38.6}
\\
\bottomrule
\end{tabular}
}
\label{tab:gcg}
\end{minipage}
\vspace{-6pt}
\end{table*}

\begin{table*}[ht]
  \centering
\caption{Unleashing visual chain-of-thought reasoning with both excellent grounding and instruction-following ability.}
\vspace{-6pt}
\scalebox{0.88}
{\begin{tabular}{l|c|cccccc|cc}
  \toprule
 \multirow{2}{*}{Model} & Visual& \multicolumn{6}{c|}{VisCoT Benchmark}& \multicolumn{2}{c}{POPE}\\
&CoT& DocVQA&TextCaps&TextVQA&DUDE &SROIE&Infographics &Acc&F1\\ \hline
 VisCoT-7B~\cite{shao2024visualcot}&\cmark&47.6&67.5&77.5&38.6&47.0&32.4&86.5&-\\ \hline
F-LMM &\xmark&43.2&63.5&74.5&32.0&28.4&43.2&87.0&86.0\\
F-LMM &\cmark&53.8&67.9&78.4&42.3&44.1&49.1&88.0&87.7\\
\bottomrule
\end{tabular}

}
\label{tab:visual_cot}
\end{table*}

\noindent\textbf{Mask Refiner.}
The mask refiner $f_r$ is retrofitted from the mask head of SAM~\cite{kirillov2023sam}, which predicts masks based on prompts as well as image embeddings from SAM's ViT-based image encoder. To refine the output of the mask decoder $f_d$, we re-use SAM's prompt encoder to transform $\rmM_{\mathrm{logits}}$ into dense prompt embeddings (\ie, a 2-D feature map) and the bounding box of $\rmM_{\mathrm{pred}}$ to box embeddings. In addition to the location cues from the mask and the box, the language cues, \ie, the object's corresponding text embeddings, are also utilised by $f_r$. Considering text embeddings from the $M$ transformer layers, we train $M$ learnable scalars to calculate a weighted sum of these text embeddings. The weighted-summed text embeddings are processed by a linear layer and then concatenated with the box embeddings to form sparse prompt embeddings. The dense and sparse prompt embeddings, together with SAM's image embeddings, are fed to the mask refiner $f_r$ for finer-grained mask predictions $\rmM'_{\mathrm{pred}}$. During training, we keep the ViT-based image encoder of SAM frozen and optimise the mask refiner $f_r$ using BCE loss and DICE loss~\cite{sudre2017dice}. For more details on the SAM's prompt-based mask head, please refer to the original SAM paper~\cite{kirillov2023sam}.

\subsection{Keyword Selector for Grounded Conversation}\label{sec:grounding_target_selection}

In grounded conversation~\cite{hanoona2023GLaMM} with interleaved segmentation masks and words, existing grounding LMMs~\cite{hanoona2023GLaMM, ren2023pixellm} expand LLMs' vocabularies with special tokens that indicate the start and end of grounding targets, which is infeasible in F-LMM given its `frozen` nature. A common practice to discover visual objects in text sequences offline is using external tools such as SpaCy~\cite{spacy2}, which parse all nouns from a sentence including unwanted non-object words. Instead of adopting such offline tools that produce noisy results, we automate generating interleaved words and masks by training a linear layer to directly predict whether a word is to be grounded or not. 

Specifically, the linear layer (keyword selector) is placed on top of the LLM's transformer layers, projecting $d$-dimension hidden state vectors into one-dimension scores, followed by a sigmoid function that normalizes the scores to $[0, 1]$. During training, the score prediction is supervised by a BCE loss. During inference, word tokens with scores exceeding a threshold $\lambda$ are regarded as positive for visual grounding. In practice, we set $\lambda = 0.3$. Adjacent positive word tokens are grouped to indicate a single visual object. After the word tokens of visual objects are selected, the corresponding attention maps and text embeddings are fed to the mask head for visual grounding.

\section{Experiments}

\subsection{Implementation Details}
\label{sec:imple}

\noindent\textbf{Model Architectures.} We build F-LMM on several open-sourced LMMs, including LLaVA-1.5~\cite{liu2023improvedllava}, LLaVA-Next~\cite{liu2024llavanext}, MiniGemini~\cite{li2024mgm}, DeepSeekVL~\cite{lu2024deepseekvl} and HPT-Air~\cite{hpt}. The main experiment covers 10 LMMs with model sizes ranging from 1.3B to 8B.
We employ a lightweight 3-stage U-Net~\cite{ronneberger2015u} as the mask decoder to transform segmentation priors from frozen LLMs.
The U-Net architecture features an encoder-decoder structure with skip connections, wherein feature maps are downsampled in the encoder and upsampled in the decoder. Please check the supplementary material for more details on the mask decoder.
As for the SAM-based mask refiner,
we choose SAM ViT-L~\cite{kirillov2023sam} that balances cost and performance well.

\noindent\textbf{Model Training.} We train F-LMM on RefCOCO(+/g)~\cite{refcoco, refcocog} and PNG~\cite{png} datasets with about 190k data samples on a single machine with 8 NVIDIA A800-40G GPUs, which costs about 20 hours for each round of model training. We set the batch size to 8 and train models for 8 epochs, with gradient clipping at a max norm of 1.0. The AdamW~\cite{loshchilov2019decoupled} optimiser is used with a learning rate of 1e-4, a weight decay of 0.01, and betas as (0.9, 0.999). We choose a warm-up ratio of 0.03 to stabilise model optimisation.

\subsection{Standard QA and Grounding Tasks}
For a comprehensive study of LMMs' conversational and grounding capabilities, we first evaluate models under standard question-answering and grounding benchmarks separately. We summarise the evaluation results of grounding LMMs in Table~\ref{tab:benchmarks}. Please refer to the supplementary material for more detailed results.

\noindent\textbf{Benchmarks.}
For comprehensive \emph{conversational ability} evaluation, we choose four widely used general question-answering benchmarks including MME~\cite{fu2023mme}, MMBench~\cite{liu2024mmbench}, LLaVA-In-the-Wild~\cite{liu2023llava} and MMVet~\cite{yu2024mmvet}. The MME and MMBench require an LMM to strictly follow the instruction to reply with single words (yes or no) or answer MCQs with alphabetical letters (\ie, answering A, B, C, or D). The LLaVA-In-the-Wild and MMVet benchmarks ask a model to respond with open-ended texts while demanding general world knowledge comprehension. In terms of \emph{grounding ability} evaluation, we assess the LMMs' ability to segment user-described objects on referring expression segmentation (RES)~\cite{refcoco, refcocog} benchmarks including RefCOCO, RefCOCO+, and RefCOCOg, using the cIoU metric.
Due to limited space, we only report results on the \texttt{Val} splits of RefCOCO(+/g) in Table~\ref{tab:benchmarks}.
We also test the LMMs' ability to ground key phrases or words in user-model conversations on the Panoptic Narrative Grounding (PNG)~\cite{png} benchmark, measuring individual mask recalls on thing/stuff objects and overall recall scores.

\noindent\textbf{Comparisons with Existing Methods.} We compare F-LMM with existing grounding LMMs. As shown in Table~\ref{tab:benchmarks}, our F-LMM provides the best balance with conversational and grounding abilities among compared methods. 
On the question-answering benchmarks, existing grounding LMMs obtain zero or near-zero scores on MMBench and MME while lagging significantly behind general-purpose LMMs on MMVet and LLaVA-In-the-Wild benchmarks, indicating compromised instruction-following ability and weakened general knowledge comprehension. On the RES and PNG benchmarks, our F-LMM achieves comparable results despite not having the parameters of LMMs fine-tuned for grounding purposes.

\begin{figure}[t]
  \centering
\includegraphics[width=0.48\textwidth]{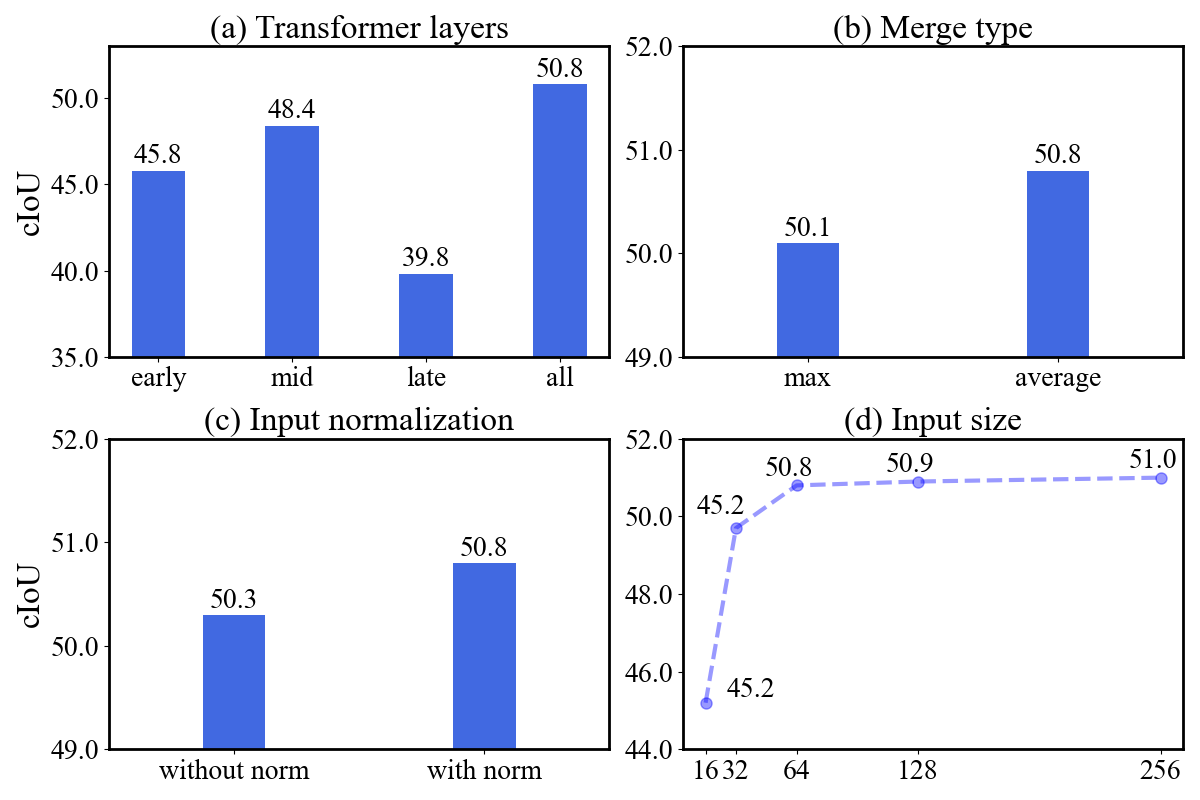}
 \vspace{-16pt}
  \caption{Ablation study of the mask decoder.}
  \label{fig:mask_decoder_alation}
   \vspace{-8pt}
\end{figure}

\subsection{Complex Scenarios}
In this section, we evaluate grounding LMMs under more complex scenarios that typically require the LMMs to perform both reasoning and segmentation. The base LMM we use is DeepSeekVL-7B~\cite{lu2024deepseekvl}, considering its flexibility in supporting both single and multiple image inputs. And the size of models compared in this section is also 7B.

\noindent\textbf{Reasoning Segmentation} is proposed in LISA~\cite{lai2023lisa} that requires a model to infer what object to segment from common-sense knowledge or via logical reasoning. The evaluation results are provided in Table~\ref{tab:reason_seg} and the metric is gIoU. F-LMM can effectively perform reasoning segmentation even though it is not trained on such type of data. It is remarkable that F-LMM significantly outperforms existing models on the subset of long sentences, reflecting the advantage of F-LMM in handling complex contexts.

\noindent\textbf{Grounded Conversation Generation} (GCG) is proposed in GLaMM~\cite{hanoona2023GLaMM}, which requires a model to generate interleaved segmentation masks and texts. For performance evaluation, METEOR (M.) and mIoU are used to measure the quality of generated texts and masks, respectively. In addition, we report the recall of object masks. As shown in Table~\ref{tab:gcg}, F-LMM exhibits the best zero-shot performance while being comparable with models fine-tuned on GCG training dataset.

\begin{table}[t]
\centering
\caption{Ablation study of the mask refiner on PNG benchmark.}
\vspace{-8pt}
\begin{minipage}[t]{0.22\textwidth}
\centering
\subcaption{Effects of Mask Refiner}
\scalebox{0.60}{\begin{tabular}{ccc|ccc}
  \toprule
\multicolumn{3}{c|}{Mask Refiner} &\multicolumn{3}{c}{PNG} \\
mask&box&text&All&Thing&Stuff \\ \hline
\xmark&\xmark& \xmark&50.8&48.6&55.9\\
\cmark &\xmark&\xmark&63.4&62.0&66.8\\
\cmark &\cmark&\xmark&63.7&62.2&67.1\\
\cmark&\cmark&\cmark &64.9&63.4&68.3\\
\bottomrule
\end{tabular}}
\label{tab:ablation_prompts}
\end{minipage}
\label{tab:ablation}
\begin{minipage}[t]{0.22\textwidth}
\centering
\subcaption{SAM Variants}
\scalebox{0.64}{
\begin{tabular}{l|ccc}
  \toprule
  \multirow{2}{*}{SAM} &\multicolumn{3}{c}{PNG} \\
  &All&Thing&Stuff \\ \hline
 ViT-B &63.0&61.4&66.8\\
 ViT-L &64.9&63.4&68.3\\
ViT-H &65.0&63.5&68.3\\
\bottomrule
\end{tabular}}
\label{tab:ablation_sam}
\end{minipage}
\vspace{-4pt}
\end{table}

\begin{table}[t]
  \centering
\caption{Ablation study of keyword selection.}
\vspace{-8pt}
\scalebox{0.75}
{




\begin{tabular}{l|ccc}
  \toprule
 Method & F1 &Recall &Precision \\ \hline
 SpaCy Parser~\cite{spacy2}&57.8 &97.3&41.1\\

 Linear Keyword Selector &82.8&96.6&72.5\\


\bottomrule
\end{tabular}
}
\vspace{-4pt}
\label{tab:keywords}
\end{table}

\begin{figure*}[t]
  \begin{minipage}[b]{0.45\linewidth}
    \centering
        \scalebox{0.93}{
\begin{tabular}{l|c|c}
  \toprule
Model & Chat $\downarrow$ & Ground $\downarrow$ \\\hline
DeepseekVL-1.3B~\cite{lu2024deepseekvl}&7.75&8.33
\\
MGM-2B~\cite{li2024mgm}&6.00& 8.33\\ \hline
LLaVA-1.5-7B~\cite{liu2023improvedllava}&6.75& 7.83\\
HPT-Air-6B~\cite{hpt}&9.00& 7.16\\ 
HPT-Air-1.5-8B~\cite{hpt}&6.50&7.00\\ 
MGM-7B~\cite{li2024mgm}&5.75&4.83\\
DeepseekVL-7B~\cite{lu2024deepseekvl}&3.75&4.00
\\\hline
LLaVA-1.6-7B~\cite{liu2024llavanext}&2.75&3.00\\
LLaVA-1.6-M-7B~\cite{liu2024llavanext}&3.25&1.66\\
MGM-HD-7B~\cite{li2024mgm}&3.50&2.83\\
\bottomrule
\end{tabular}

}
  \end{minipage}
  \hspace{-16pt}
  \begin{minipage}[b]{.58\linewidth}
    \centering
    \includegraphics[width=0.85\linewidth]{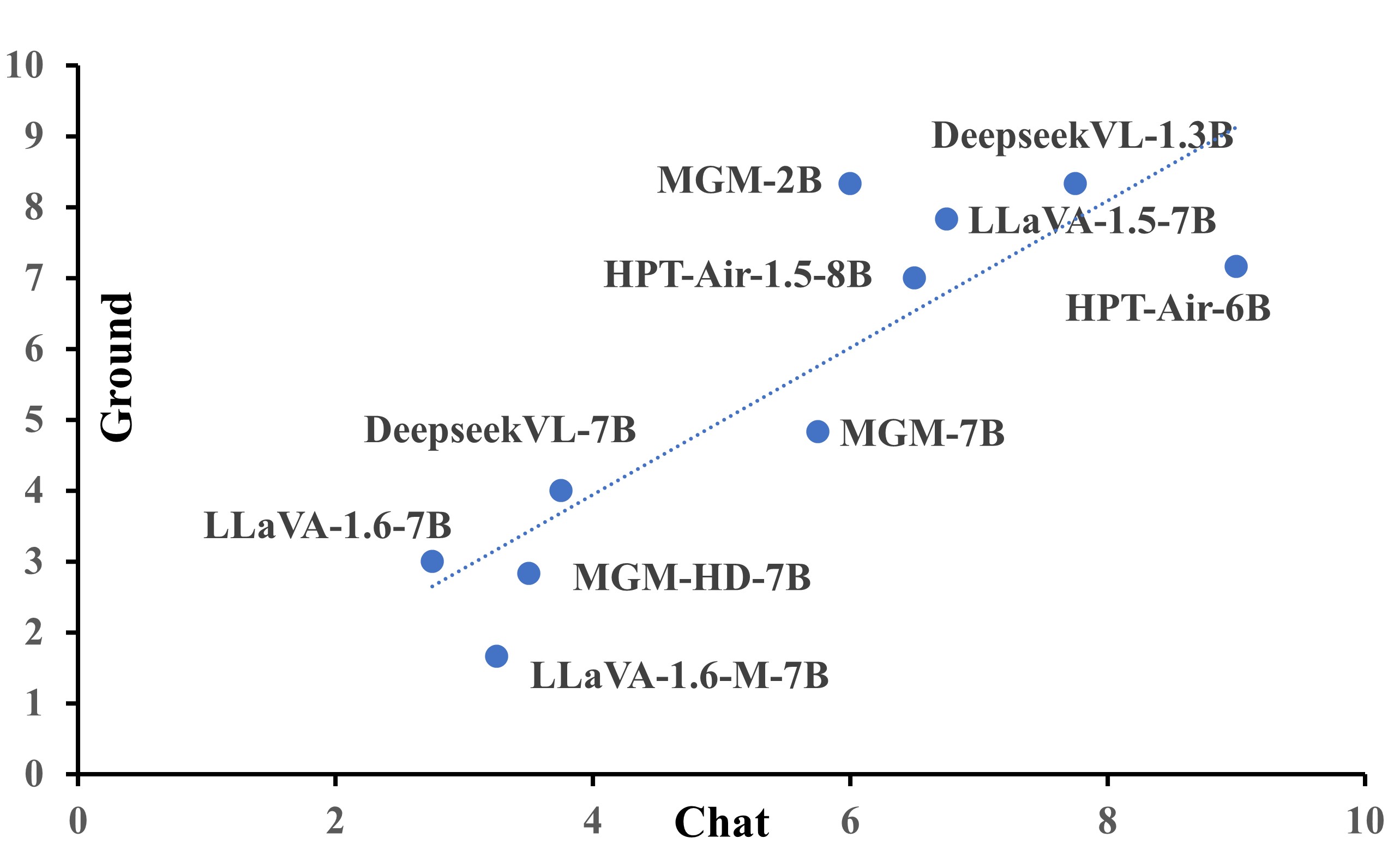}
        \vspace{-70pt}
  \end{minipage}
        \captionof{figure}{
The relevance between an LMM's chat ability and the grounding ability of the F-LMM built on it.
        The left table shows the average ranks of each LMM on question-answering (`Chat') and grounding benchmarks (`Ground'). $\downarrow$ means the lower the better. The dashed line in the right figure is the linear fit of the rank data points, indicating a positive correlation between abilities to chat and ground.}
\label{fig:ranks}
\vspace{-4pt}
\end{figure*}

\noindent\textbf{Visual Chain-of-Thought Reasoning.} In human-AI conversations that involve Visual Chain-of-Thought Reasoning (VisCoT)~\cite{shao2024visualcot}, an LMM first localises the region/object relevant to the human's question and then generates the final answer by zooming in on the question-related region. Here we evaluate F-LMM and VisCoT-7B on the VisCoT benchmark~\cite{shao2024visualcot}. 
As shown in Table~\ref{tab:visual_cot}, F-LMM achieves remarkable performance gains when prompted in the VisCoT manner. It is noticeable that F-LMM even outperforms VisCoT-7B~\cite{shao2024visualcot} that has been well-tuned on the VisCoT training data~\cite{shao2024visualcot}. Furthermore, we perform VisCoT on the object hallucination benchmark POPE~\cite{li2023pope} and observe significant performance gain in resisting object hallucinations. 

\subsection{Ablation Study}
We investigate the effects of design choices of F-LMM. All the ablation studies are conducted on the PNG dataset and we use DeepSeekVL-1.3B~\cite{lu2024deepseekvl} unless otherwise stated.

\noindent\textbf{Mask Decoder.} We summarise our analyses of the mask decoder in Figure~\ref{fig:mask_decoder_alation}. Note that the mask refiner is not involved in this part. We first consider attention maps from different transformer layers, \ie, early ($6^{\text{th}}$), mid ($12^{\text{th}}$) and late ($24^{\text{th}}$) layers. As shown in Figure~\ref{fig:mask_decoder_alation}(a), attention maps from late layers perform the worst, conforming to prior studies~\cite{huang2024opera} indicating that deeper transformer layers tend to focus on abstract concepts instead of visual details. And using attention maps from all layers achieves the best performance. Next, we study how to merge the word-image attention maps of multi-word objects, as shown in Figure~\ref{fig:mask_decoder_alation}(b). We find that the average operation outperforms the max operation by a margin of 0.7. In 
Figure~\ref{fig:mask_decoder_alation}(c), we show that normalizing the inputs to the mask decoder provides a 0.5 performance gain. Finally, we experiment with different input sizes for the mask decoder. As shown in Figure~\ref{fig:mask_decoder_alation}(d), using $64\times 64$ yields the best performance-cost trade-off.

\noindent\textbf{Mask Refiner.} We study the effects of the mask refiner on segmentation in Table~\ref{tab:ablation_prompts}. The performance of using only the mask decoder is shown in the first row of Table~\ref{tab:ablation_prompts}.
With the masks fed to the mask refiner, we observe a significant 12.6 performance gain on the PNG benchmark. Adding box and text prompts to the mask refiner further improves the performance by 0.3 and 1.2. Then we experiment with different SAM~\cite{kirillov2023sam} model variants, \ie, ViT-B(ase), ViT-L(arge) and ViT-H(uge) as mask refiners. As shown in Table~\ref{tab:ablation_sam}, the performance grows with model sizes. For a good trade-off between cost and performance, we select the ViT-L variant of SAM as the default mask refiner. 

\noindent\textbf{Keyword Selector.} We analyse the keyword selector, implemented as a linear layer, on the PNG dataset, using the F1 score as the main metric. As shown in Table~\ref{tab:keywords}, our keyword selector achieves significantly higher F1 scores than the external SpaCy tool~\cite{spacy2}. We also report recall and precision scores. Our keyword selector achieves comparable recall while being much more precise compared to SpaCy, which enumerates all nouns in a sentence.

\subsection{Analysis \& Visualisation}

\noindent\textbf{Scalability: Does Better Chatting Lead to Better Grounding?} We study the relevance between an LMM's chat ability and the grounding ability of the F-LMM built on it. Specifically, we examine the correlation between performance on the question-answering and grounding benchmarks. For the ten models reported in Table~\ref{tab:benchmarks}, we calculate their average ranks in each benchmark category. In Figure~\ref{fig:ranks}, we plot these ranks as 2D coordinates, \ie, (\texttt{Chat Rank}, \texttt{Ground Rank}), and apply a linear fit to the data points. As indicated by the blue dashed line, frozen LMMs with stronger conversational ability can serve as better backbones for grounding. We also observe that larger LMMs generally excel in both conversation and grounding tasks, and LMMs with larger input resolutions (\eg, LLaVA-1.6 and MGM-HD) can handle both tasks better.

\begin{figure}[t]
  \centering
\includegraphics[width=0.45\textwidth]{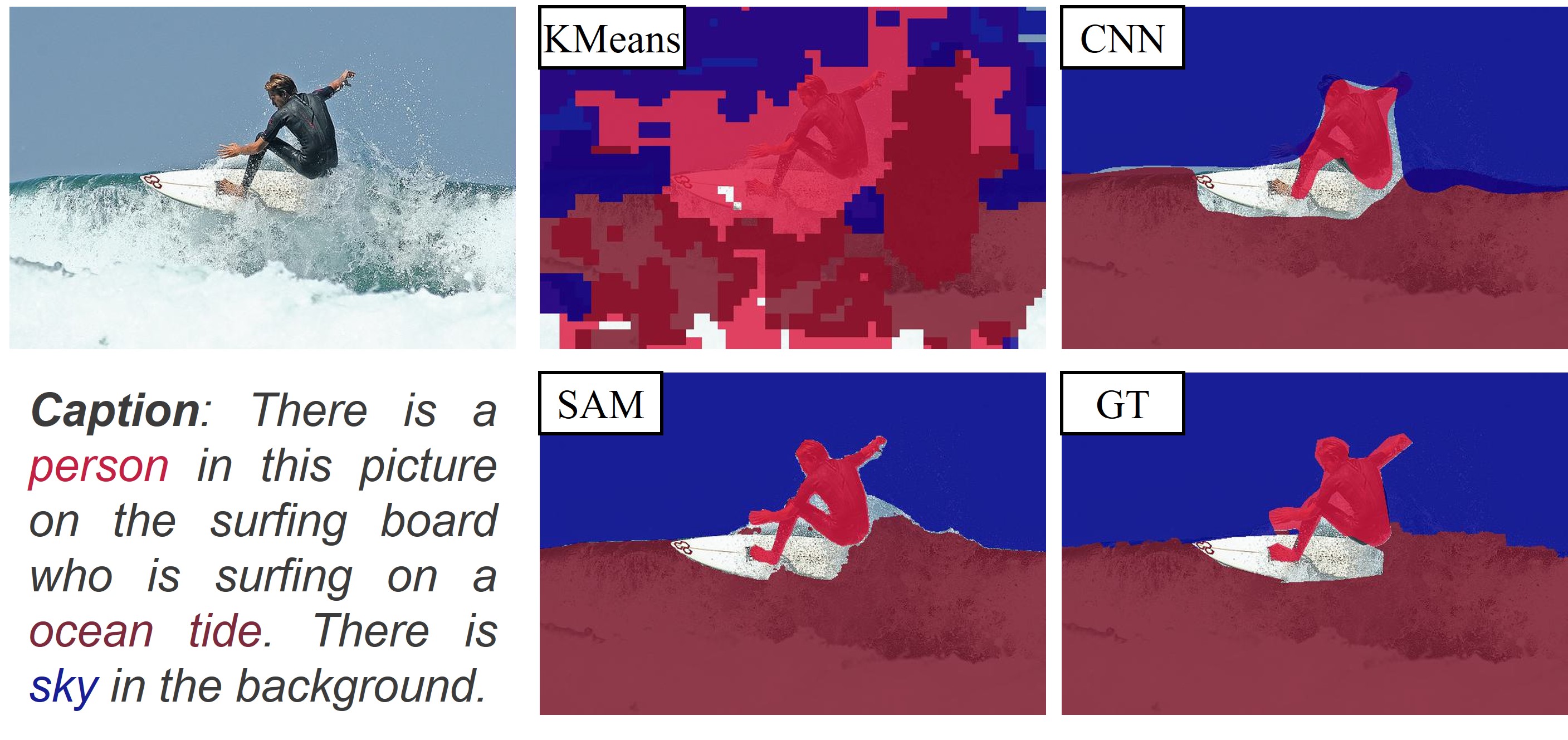}
\vspace{-12pt}
  \caption{Visualisations of KMeans clustering on attention maps and segmentation results from the CNN-based mask decoder and the SAM-based mask refiner.
  }
  \vspace{-10pt}
  \label{fig:png_kmeans_vis}
\end{figure}

\noindent\textbf{From Attention Maps to Segmentation Masks.}
We visualise the word-image attention maps by applying KMeans clustering to the stacked attention maps that are collected from all transformer layers and attention heads. The attention maps of multiple-word objects are merged by element-wise average. As shown in Figure~\ref{fig:png_kmeans_vis}, we observe that the pixels of objects are roughly clustered together (top-left). With the CNN-Based mask decoder, the attention weights are mapped to 2D binary masks (top-right), which are then further optimised by the SAM-based mask refiner (bottom-left). The model used is DeepSeekVL-1.3B~\cite{lu2024deepseekvl}.

\section{Conclusion}
In this work, we have studied the limitation of existing grounding LMMs, \ie, the loss of general world knowledge and instruction-following ability. To address this issue, we make the first attempt to ground fully frozen LMMs, which are already well-trained for user-model conversation, based on the insight that the geometric and spatial cues needed for visual grounding are inherently present within the self-attention mechanism of LMMs. By incorporating a CNN-based mask decoder and a SAM-based mask refiner, we achieve competitive visual grounding performance without sacrificing any conversational abilities of pre-trained LMMs. With this combination of strong conversational and visual grounding capabilities, these LMMs show promise for complex perception and reasoning tasks, such as segmentation with reasoning, grounded conversation generation, and visual chain-of-thought reasoning.

\setcounter{table}{0}
\renewcommand{\thetable}{A\arabic{table}}
\setcounter{figure}{0}
\renewcommand{\thefigure}{A\arabic{figure}}
\setcounter{section}{0}
\renewcommand{\thesection}{A\arabic{section}}
\section{Appendix}
In Sec~\ref{sec:sup_benchmarks}, we provide more detailed experimental results on both question-answering benchmarks and grounding benchmarks. Then we introduce the architecture of F-LMM's mask decoder in Sec~\ref{sec:sup_architecture}.
We also summarise the datasets used by F-LMM and existing grounding LMMs in Sec~\ref{sec:sup_datasets}.
Last, we provide visualisation results in Sec~\ref{sec:sup_visualization}, including failure cases of existing grounding LMMs on general question-answering tasks and examples of reasoning segmentation, grounded conversations and visual CoT. The broader impact and limitations of this work are elucidated in Sec~\ref{sec:impact} and Sec~\ref{sec:limitation}, respectively.

\subsection{Benchmark results}\label{sec:sup_benchmarks}

\noindent\textbf{Question-Answering Benchmarks.}
In addition to the four benchmarks reported in the main text, we test the grounding LMMs on a wider range of question-answering benchmarks as shown in Table~\ref{tab:sup_qa_benchmarks}. Due to corrupted instruction-following abilities, existing grounding LMMs obtain zero or near-zero scores on these benchmarks.

\begin{table*}[h]
  \centering
\caption{More evaluation results on question-answering benchmarks.}
\scalebox{0.9}
{\begin{tabular}{l|cccccccc}
  \toprule
Model&MME&MMBench&MMVet&LLaVA$^{\mathrm{W}}$&POPE&GQA&VQA$^{\mathrm{v2}}$&AI2D\\ \hline
\multicolumn{9}{c}{\textit{Existing Grounding LMMs}} \\\hline
PixelLM-7B~\cite{ren2023pixellm}&309/135&17.4&15.9&46.4&0.0&0.0&0.0&0.0\\
PixelLM-13B~\cite{ren2023pixellm}&77/47 & 18.1 & 18.1 & 47.8&0.0&0.0&0.0&0.0\\
LISA-7B~\cite{lai2023lisa}&1/1&0.4&19.1&47.5&0.0&0.0&0.0&0.0\\
LISA-13B~\cite{lai2023lisa}&2/1&0.8&19.8&48.1&0.0&0.0&0.0&0.0\\
LLaVA-G-7B~\cite{zhang2023llavagrounding}&-&-&-&55.8&-&-&-&-\\
GLaMM-7B~\cite{hanoona2023GLaMM}&14/9&36.8&10.3&32.0&0.94&11.7&24.4&28.2\\
LaSagnA-7B~\cite{wei2024lasagna}&0/0&0.0&16.7&34.5&0.0&0.0&0.0&0.0\\ \hline
\multicolumn{9}{c}{\textit{General-Purpose LMMs}} \\\hline
DeepseekVL-1.3B~\cite{lu2024deepseekvl}&1307/225&64.6&34.8&51.1&88.3&59.3&76.2&51.5
\\
MGM-2B~\cite{li2024mgm}&1341/312&59.8 &31.1&65.9&83.9&59.9&72.9&62.1\\ \hline
LLaVA-1.5-7B~\cite{liu2023improvedllava}&1511/348&64.3&30.5&69.0&85.9&62.0&76.6&54.8\\
HPT-Air-6B~\cite{hpt}&1010/
258&69.8&31.3&59.2&87.8&56.2&74.3&64.8\\ 
HPT-Air-1.5-8B~\cite{hpt}&1476/308&75.2&36.3&62.1&90.1&59.4&78.3&69.0\\ 
MGM-7B~\cite{li2024mgm}&1523/316&69.3&40.8&75.8&84.2&61.6&76.7&64.3\\
DeepseekVL-7B~\cite{lu2024deepseekvl}&1468/298&73.2&41.5&77.8&88.0&61.3&78.6&65.3
\\\hline
LLaVA-1.6-7B~\cite{liu2024llavanext}&1519/322&68.1&44.1&72.3&86.4&64.2&80.2&66.6\\
LLaVA-1.6-Mistral-7B~\cite{liu2024llavanext}&1501/324&69.5&47.8&71.7&86.8&55.0&80.3&60.8\\
MGM-HD-7B~\cite{li2024mgm}&1546/319&65.8&41.3&74.0&84.2&61.6&76.7&64.3\\
\bottomrule
\end{tabular}
}
\label{tab:sup_qa_benchmarks}
\end{table*}

\noindent\textbf{Referring Expression Segmentation.} The results reported in the main text only include scores on the \texttt{Val} subsets of RefCOCO, RefCOCO+ and RefCOCOg. Here, we provide the grounding LMMs' performances on all their subsets in Table~\ref{tab:sup_refcoco}. The metric used for Referring Expression Segmentation (RES) is cIoU.

\begin{table*}[h]
  \centering
\caption{Detailed comparisons on Referring Expression Segmentation (RES).}
\scalebox{0.9}
{\begin{tabular}{l|ccc|ccc|cc}
  \toprule
 \multirow{2}{*}{Model}&\multicolumn{3}{c|}{RefCOCO} &\multicolumn{3}{c|}{RefCOCO+}&\multicolumn{2}{c}{RefCOCOg}  \\ 
&val&testA&testB&val&testA&testB&val&test\\ \hline

\multicolumn{9}{c}{\textit{Specialised Segmentation Models}} \\\hline
MCN~\cite{luo2020mcn} &62.4 &64.2 &59.7 &50.6 &55.0 &44.7 &49.2& 49.4\\
LAVT~\cite{yang2022lavt} & 72.7 &75.8 &68.8 &62.1 &68.4&55.1&61.2 &62.1
\\
GRES~\cite{liu2023gres} &73.8 &76.5 &70.2 &66.0&71.0 &57.7 &65.0 &66.0\\
X-Decoder~\cite{zou2023generalized} & -& - &- &- &- &- &64.6& -\\
SEEM~\cite{zou2024seem} & -& - &- &- &- &- & 65.7&-\\
\hline
\multicolumn{9}{c}{\textit{Existing Grounding LMMs}} \\\hline
PixelLM-7B~\cite{ren2023pixellm}&73.0&76.5&68.2& 66.3&71.7&58.3&69.3& 70.5\\
LISA-7B~\cite{lai2023lisa}&74.9 &79.1 &72.3 &65.1 &70.8 &58.1 &67.9 &70.6\\
PerceptionGPT-7B~\cite{pi2023perceptiongpt}&75.1 &78.6 &71.7 &68.5 &73.9 &61.3&70.3&71.7\\
LLaVA-G-7B~\cite{zhang2023llavagrounding}&77.1&-&-&68.8&-&-&71.5&-\\
GroundHog-7B~\cite{zhang2024groundhog}&78.5&79.9&
75.7&70.5&75.0&64.9&74.1&74.6\\
GLaMM-7B~\cite{hanoona2023GLaMM}&78.6&81.1&76.1&70.5&74.9&63.0&74.8&74.8\\
LaSagnA-7B~\cite{wei2024lasagna}&76.8&78.7&73.8&66.4&70.6&60.1&70.6& 71.9\\ \hline
\multicolumn{9}{c}{\textit{Grounding Frozen General-Purpose LMMs by F-LMM (Ours)}} \\
\hline
DeepseekVL-1.3B~\cite{lu2024deepseekvl}&75.0&78.1&69.5&62.8&70.8&56.3&68.2&68.5\\
MGM-2B~\cite{li2024mgm}&75.0&78.6&69.3&63.7&71.4&53.3&67.3&67.4\\ 
\hline
LLaVA-1.5-7B~\cite{liu2023improvedllava}&75.2&79.1&71.9&63.7&71.8&54.7&67.1&68.1
\\
HPT-Air-6B~\cite{hpt}&74.3&79.4&71.8&64.0&71.7&57.2&67.5&68.3\\ 
HPT-Air-1.5-8B~\cite{hpt}&76.3&78.5&70.8&64.5&72.8&55.4&68.5&69.6\\ 
MGM-7B~\cite{li2024mgm}&75.7&80.2&70.8&64.8&73.2&55.3&68.3&69.4\\
DeepseekVL-7B~\cite{lu2024deepseekvl}&76.1&78.8&72.0&66.4&73.2&57.6&70.1&70.4\\
\hline
LLaVA-1.6-7B~\cite{liu2024llavanext}&75.8&79.5&72.4&65.8&75.2&58.5&70.1&71.7
\\
LLaVA-1.6-Mistral-7B~\cite{liu2024llavanext}&75.7&79.6&71.2&66.5&75.5&58.1&70.1&70.3
\\
MGM-HD-7B~\cite{li2024mgm}&76.1&80.2&72.0&65.2&73.4&55.7&68.5&69.4\\
\bottomrule
\end{tabular}
}
\label{tab:sup_refcoco}
\end{table*}

\noindent\textbf{Panoptic Narrative Grounding.} 
In the main text, we only report individual mask recalls on thing and stuff objects as well as the overall average recall. Here, we additionally report the mask recalls on singular and plural object nouns as shown in Table~\ref{tab:sup_png}. As expected, segmenting plural nouns that refer to multiple object instances is more challenging for all the tested models.
\begin{table*}[t]
  \centering
\caption{Detailed comparisons on Panoptic Narrative Grounding (PNG).}
\scalebox{0.9}
{\begin{tabular}{l|ccccc}
  \toprule
Model&All&Thing&Stuff&Singular&Plural\\ \hline

\multicolumn{6}{c}{\textit{Specialist Segmentation Models}} \\\hline
MCN~\cite{luo2020mcn}&54.2 &48.6 &61.4&56.6 &38.8  \\
PNG~\cite{png} &55.4&56.2&54.3&56.2&48.8 \\
PPMN~\cite{ding2022ppmn} &59.4& 57.2& 62.5&60.0&54.0 \\
XPNG~\cite{guo2024xpng} &63.3& 61.1& 66.2&64.0&56.4\\
\hline
\multicolumn{6}{c}{\textit{Existing Grounding LMMs}} \\\hline
PixelLM-7B~\cite{ren2023pixellm}&43.1&41.0&47.9&49.1&27.7\\
GroundHog-7B~\cite{zhang2024groundhog}&66.8&65.0&69.4&70.4&57.7\\
GLaMM-7B~\cite{hanoona2023GLaMM}&55.8&52.9&62.3&59.7&45.7\\
\hline
\multicolumn{6}{c}{\textit{Grounding Frozen General-Purpose LMMs by F-LMM (Ours)}} \\\hline
DeepseekVL-1.3B~\cite{lu2024deepseekvl}&64.9&63.4&68.3&68.3&
56.1\\
MGM-2B~\cite{li2024mgm}&65.6&64.4&68.4&69.1&56.9\\ \hline
LLaVA-1.5-7B~\cite{liu2023improvedllava}&64.8&63.4&68.2&68.2&56.1\\
HPT-Air-6B~\cite{hpt}&65.5&64.0&68.8&68.9&56.6\\ 
HPT-Air-1.5-8B~\cite{hpt}&65.4&64.1&68.5&68.9&56.5\\ 
MGM-7B~\cite{li2024mgm}&66.3&65.3&68.6&69.8&57.3\\
DeepseekVL-7B~\cite{lu2024deepseekvl}&65.7&64.5&68.5&69.2&56.7
\\\hline
LLaVA-1.6-7B~\cite{liu2024llavanext}&66.3&65.1&69.0&69.8&57.3\\
LLaVA-1.6-Mistral-7B~\cite{liu2024llavanext}&66.5&65.4&69.1&70.0&57.5
\\
MGM-HD-7B~\cite{li2024mgm}&66.7&65.6&69.1&70.1&57.8\\
\bottomrule
\end{tabular}
}
\label{tab:sup_png}
\end{table*}

\subsection{Mask Decoder}\label{sec:sup_architecture}
\begin{figure*}[ht]
  \centering
\includegraphics[width=0.98\textwidth]{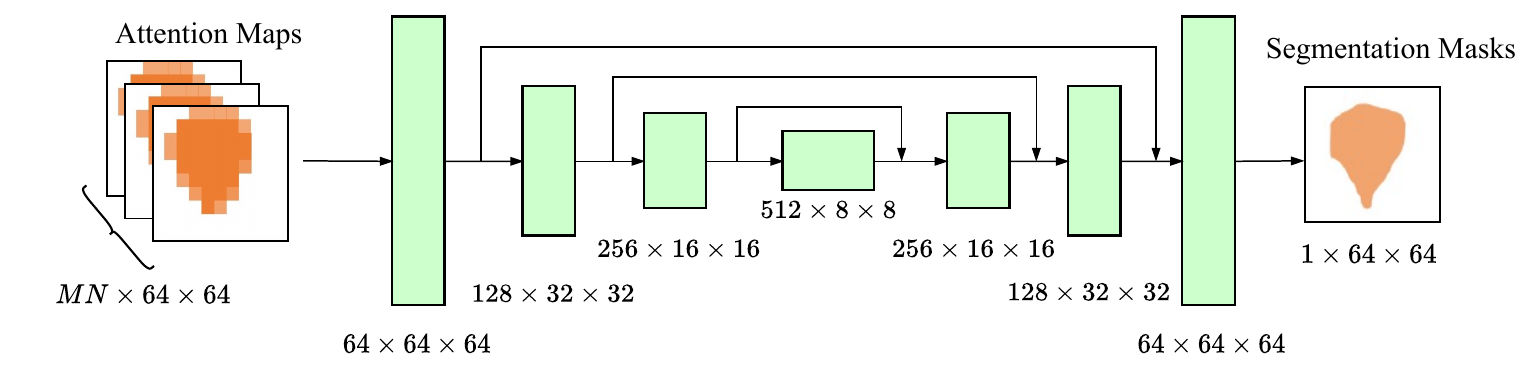}
  \caption{The architecture of the mask decoder is based on a 3-stage U-Net~\cite{ronneberger2015u} where the feature maps are downsampled and upsampled 3 times.}
  \label{fig:sup_mask_decoder}
\end{figure*}
The architecture of the mask decoder based on a 3-stage U-Net~\cite{ronneberger2015u} is shown in Figure~\ref{fig:sup_mask_decoder}, in which the feature maps are downsampled and upsampled three times. Downsampling encompasses two convolutional layers with a kernel size of 2 and 1, respectively. Upsampling is achieved using bilinear interpolation followed by two convolutional layers with a kernel size of 1. The number of parameters of the mask decoder is 8M.

\begin{table*}[h]
  \centering
\caption{Datasets used by F-LMM and existing grounding LMMs. COCO$^s$ stands for the COCO-Stuff~\cite{cocostuff} dataset and COCO$^p$ is for the COCO-Panoptic~\cite{panoptic} dataset.}
\scalebox{0.65}
{\begin{tabular}{l|ccccc|ccccccc|cccc}
  \toprule
 \multirow{2}{*}{Datasets}&\multicolumn{5}{c|}{Language-Based Segmentation} &\multicolumn{7}{c|}{Standard Segmentation}&\multicolumn{4}{c}{Grounded Conversation}  \\ 
&RefCOCO&RefCLE&VG&PNG&Flickr  &COCO$^s$&ADE&Mapillary&VLPart&COCO$^p$&Cityscapes&OpenImage&GranD&MUSE&GVC&M3G2
\\ \hline

F-LMM (Ours) &\checkmark& & &\checkmark&&&&&&&&&&& \\
LISA~\cite{lai2023lisa} & \checkmark &\checkmark &&&&\checkmark&\checkmark&\checkmark&\checkmark&&&&&&&
\\
Llava-G~\cite{zhang2023llavagrounding} &\checkmark & &\checkmark &&\checkmark &\checkmark&&&&\checkmark&&&&&\checkmark&\\
GLaMM~\cite{hanoona2023GLaMM} &\checkmark&\checkmark& \checkmark& &\checkmark&\checkmark&\checkmark&\checkmark&\checkmark&&&&\checkmark&&&\\
GroundHog~\cite{zhang2024groundhog} &\checkmark& & &\checkmark&\checkmark&&&&&&&&&&&\checkmark\\
PixelLM~\cite{ren2023pixellm} & \checkmark &\checkmark &&&&\checkmark&\checkmark&\checkmark&\checkmark&&&&&\checkmark&&\\
LaSagnA~\cite{wei2024lasagna}& \checkmark &\checkmark &&&&\checkmark&\checkmark&\checkmark&\checkmark&&\checkmark&\checkmark&&&&\\
\bottomrule
\end{tabular}
}
\label{tab:datasets}
\end{table*}

\subsection{Dataset Usage}\label{sec:sup_datasets}
\noindent\textbf{Training Data Comparison.}
In Table~\ref{tab:datasets}, we show the datasets used by existing grounding LMMs and our F-LMM. Existing methods conduct training on a wide range of standard segmentation datasets for excellent grounding ability and collect grounded conversation datasets to preserve chat ability. In contrast, F-LMM only need the RefCOCO and PNG datasets for segmentation capability, without needing additional grounded instruction-tuning datasets.

\noindent\textbf{Training Data Format.} For PNG dataset, an image narrative (\eg, `A hot air balloon is flying over the river') is formatted as \textit{`User: Describe the image. Model: A \textcolor{red}{hot air balloon} is flying over the \textcolor{cyan}{river}.'} The coloured texts indicate the keywords for grounding, which are annotated in PNG's training set. For RES dataset where each image is associated with multiple referring expressions (\eg `The man in blue T-short', `The girl who is smiling'). We convert the RES data to PNG format by concatenating the referring expressions into a single sentence: \textit{`User: Describe the image. Model: \textcolor{red}{The man in blue T-short}; \textcolor{cyan}{The girl who is smiling}.'}

\subsection{Visualisation}\label{sec:sup_visualization}
\begin{figure*}[t]
  \centering
\includegraphics[width=1.0\textwidth]{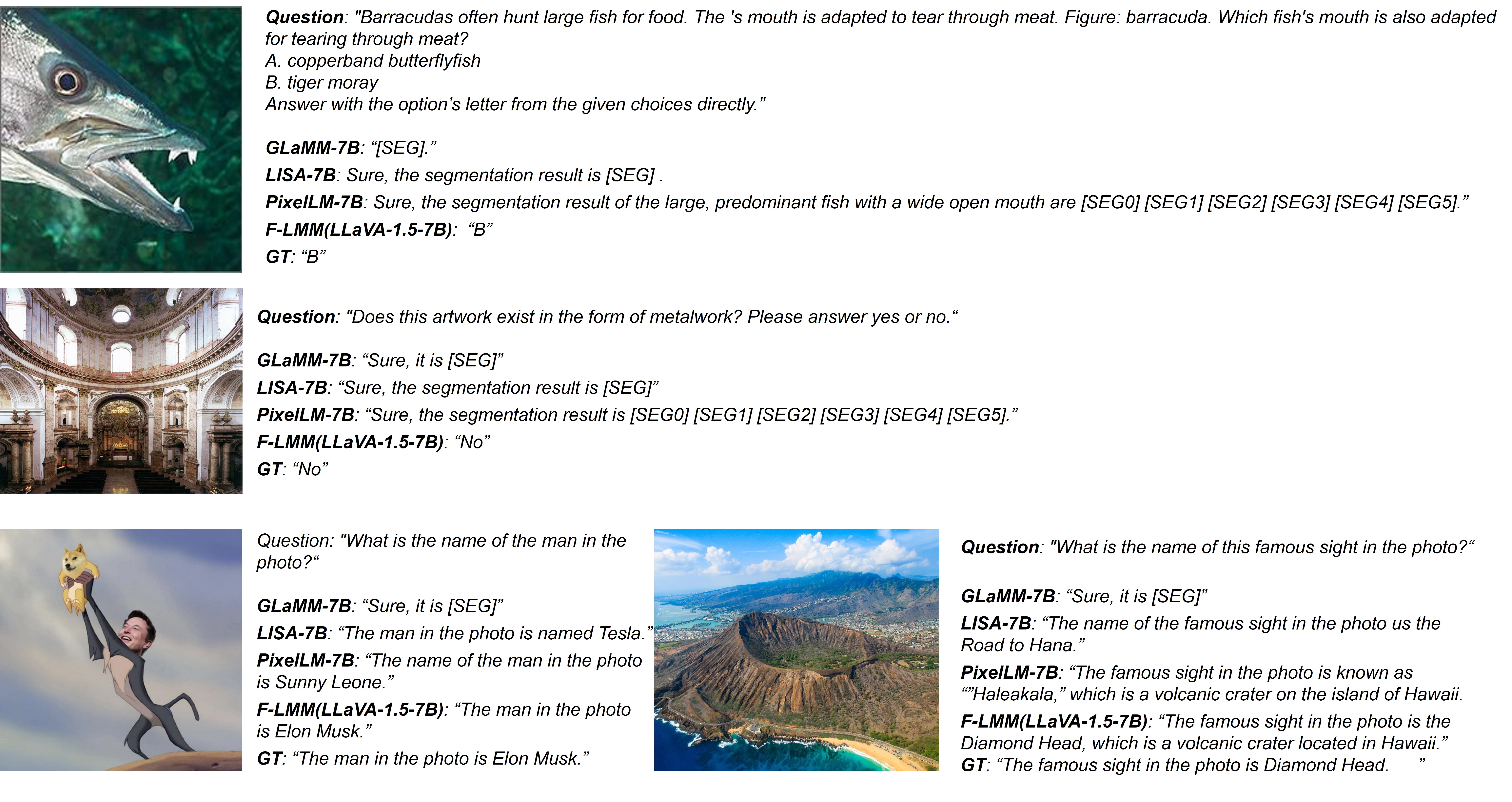}
  \caption{Examples of grounding LMMs performing general question-answering tasks. The first example is obtained from MMBench~\cite{liu2024mmbench}, the second example is extracted from MME~\cite{fu2023mme}, and the last two examples are from LLaVA-in-the-Wild~\cite{liu2023llava}. Existing grounding models (GLaMM, LISA, and PixelLM) fail to strictly follow user instructions nor correctly answer questions that necessitate a grasp of general world knowledge. In contrast, F-LMM (built upon LLaVA-1.5~\cite{liu2023improvedllava} in the above examples), which completely inherits the conversational ability of general-purpose LMMs, performs excellently on these question-answering tasks.
  }
  \label{fig:sup_more_comparisons}
\end{figure*}

\noindent\textbf{General Multimodal Question-Answering.} In Figure~\ref{fig:sup_more_comparisons}, we show some examples of grounding LMMs performing general question-answering tasks. When prompted to answer with single words (\eg, yes or no), existing grounding LMMs (GLaMM~\cite{hanoona2023GLaMM}, LISA~\cite{lai2023lisa}, and PixelLM~\cite{ren2023pixellm}) usually fail to follow the user instructions. Besides, we also observe that the grounding LMMs tend to misunderstand the user's questions as segmentation requests and reply mask tokens, \eg, `[SEG]'. Furthermore, these grounding LMMs fail to recognise the celebrity (Musk) and famous natural spot, exhibiting a worse grasp of world knowledge compared with a general-purpose LMM (\eg, LLaVA~\cite{liu2023llava}). In contrast, F-LMM inherits the virtues of general-purpose LMMs in instruction following and world knowledge comprehension, thanks to the `Frozen' design philosophy.

\begin{figure*}[t]
  \centering
    \includegraphics[width=1.0\textwidth]{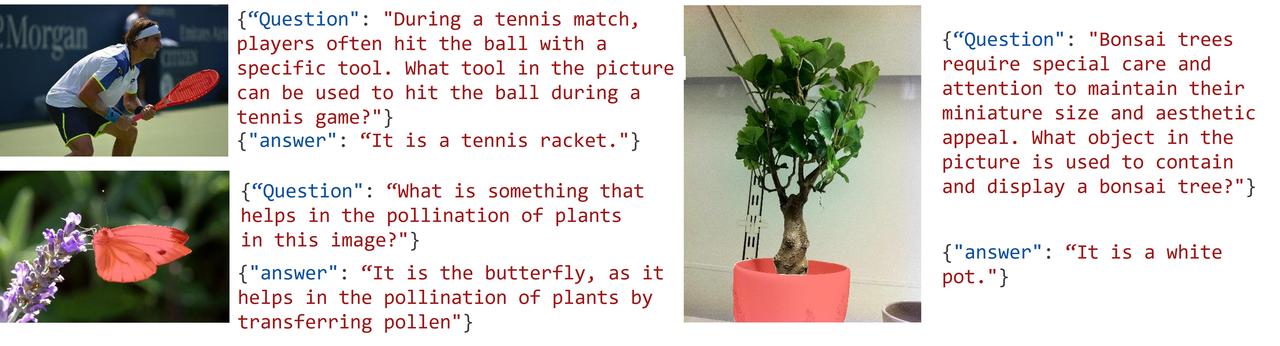}
  \caption{Examples of Reasoning Segmentation. The red masks in the images are segmentation results. The model generating the answers is DeepSeekVL-7B. The sentences in the answers are grounded by the mask head of F-LMM.}
  \label{fig:figure_r1}
\end{figure*}

\begin{figure*}[t]
  \centering
\includegraphics[width=1.0\textwidth]{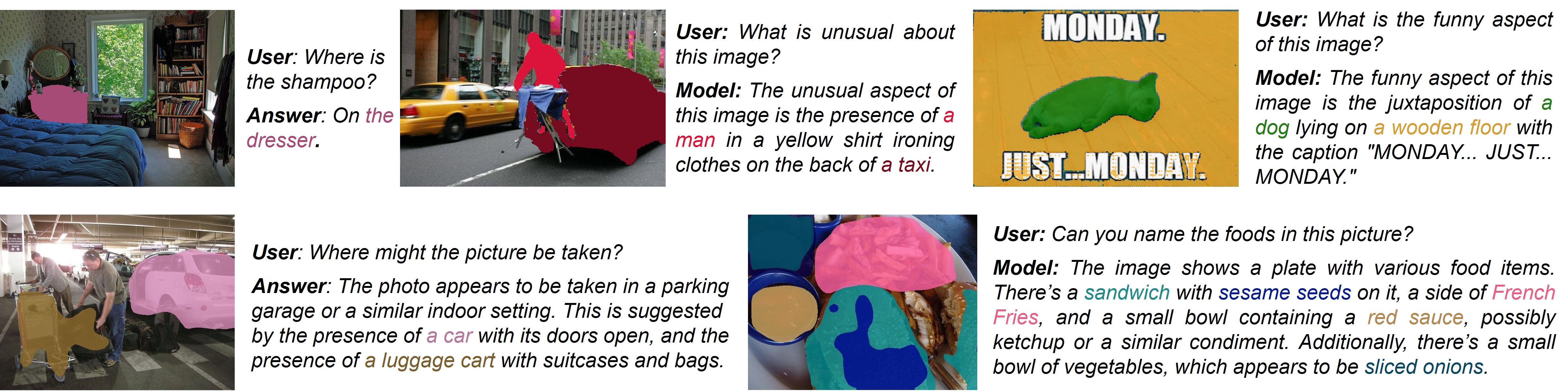}
  \caption{Visualisations of grounded human-AI conversations. The key phrases or words in the conversations can be precisely localised by the mask head of F-LMM. The LMM used is DeepSeekVL-7B.}
  \label{fig:conversation}
\end{figure*}

\begin{figure*}[t]
  \centering
\includegraphics[width=1.0\textwidth]{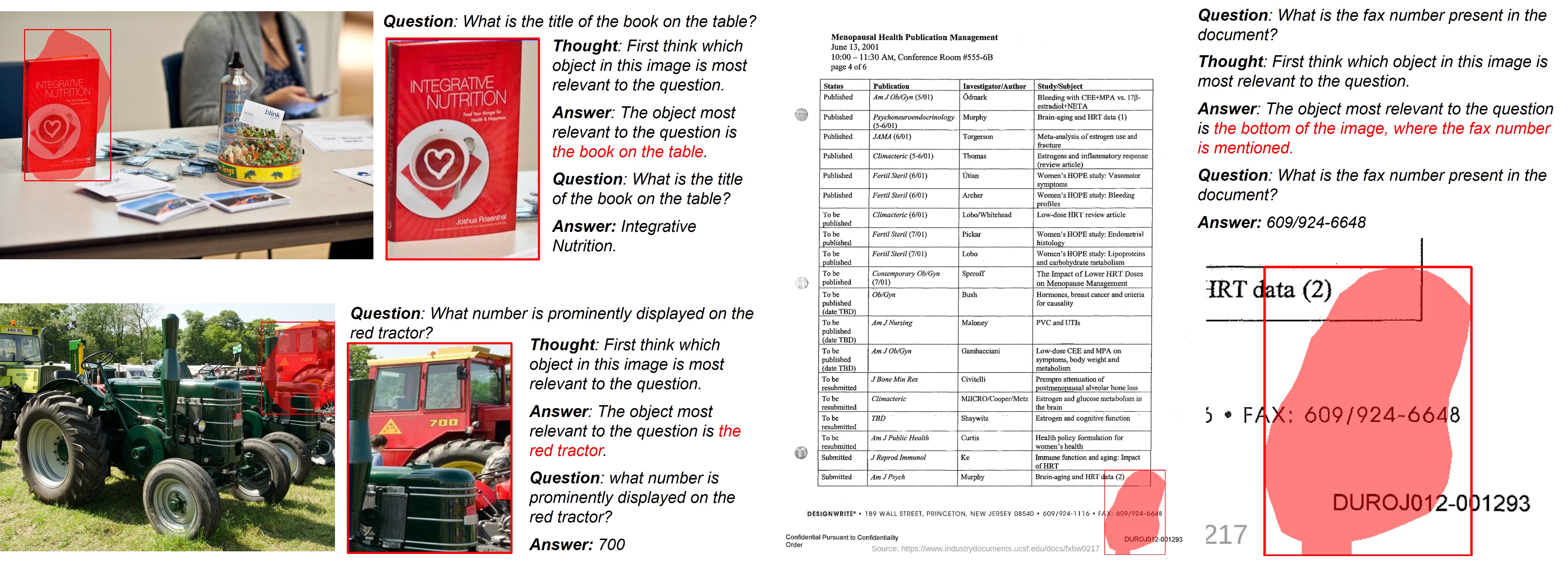}
  \caption{Visual Chain-of-Thought Reasoning. The model used is DeepSeekVL-7B, and the samples are taken from the test set of VisCoT dataset~\cite{shao2024visualcot}. The LMM is first prompted to think about the question-related object, which is then grounded by the mask head of F-LMM. The region of the question-related object is cropped and fed to the LMM to help answer the question.}
  \label{fig:sup_viscot}
\end{figure*}

\begin{figure}[t]
  \centering
\includegraphics[width=0.46\textwidth]{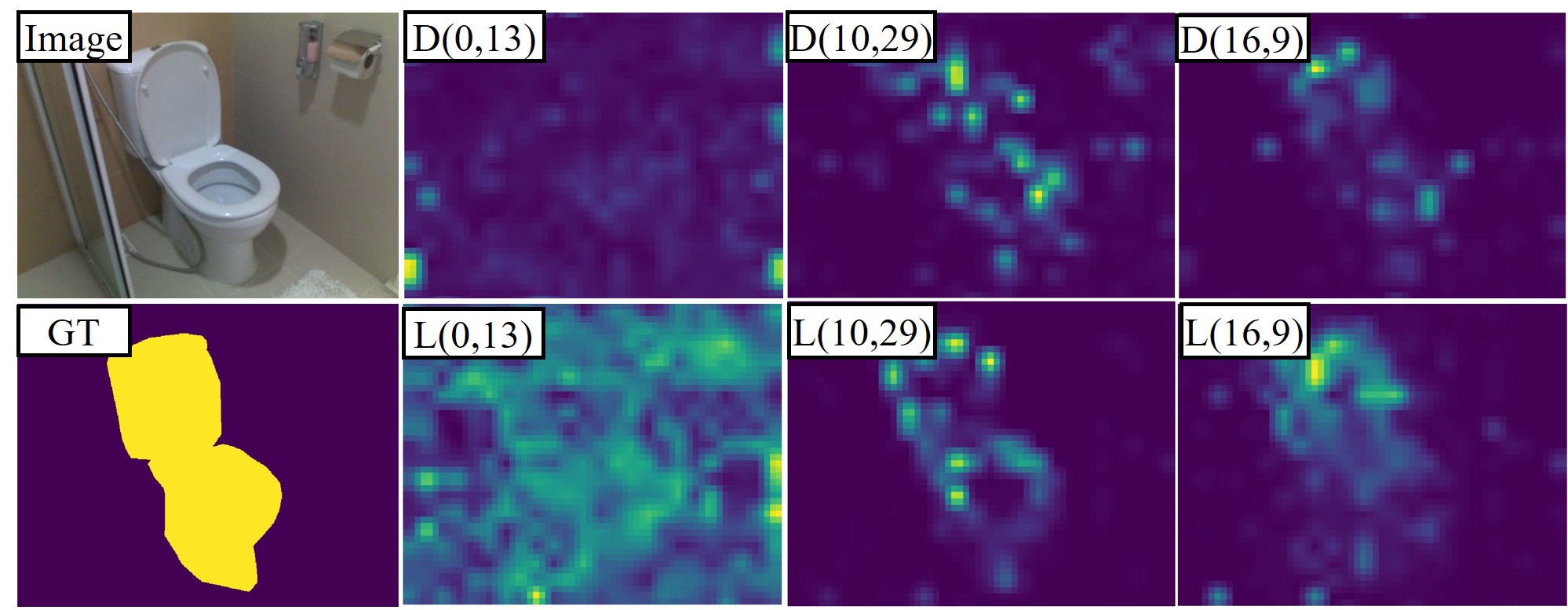}
  \caption{Comparison between the attention maps of DeepSeekVL-7B(indicated as `D') and LLaVA-1.5-7B(indicated as `L'). (m,n) means attention map at the n-th head of the m-th layer.}
  \label{fig:sup_attn_maps}
\end{figure}

\noindent\textbf{Reasoning Segmentation.} We show examples of F-LMM performing reasoning segmentation in Figure~\ref{fig:figure_r1}. The LMM is first prompted to answer a question relevant to an object in the given image. The content of the answer is regarded as the grounding target. We observe that the LMM (DeepSeekVL-7B) is able to generate the correct answer of the queried object, which is then precisely localised by the mask head of F-LMM.

\noindent\textbf{Grounded Conversation.}
In Figure~\ref{fig:conversation}, we show some examples of grounding conversation by F-LMM. Our F-LMM maintains the LMMs' original ability to follow the user's instruction and understand unusual scenarios (\eg, the man ironing at the back of a taxi) while being able to localise the keywords and phrases during the conversations precisely. The model used in these examples is DeepSeekVL-7B.

\noindent\textbf{Visual Chain-of-Thought Reasoning.}
Figure~\ref{fig:sup_viscot} shows examples of visual CoT by F-LMM. The model used in these examples is DeepSeekVL-7B. When the LMM is prompted to answer \textit{`which object is the most relevant to the question'}, the mask head of F-LMM grounds the LMM's answer about the relevant object by generating a segmentation mask, the bounding box of which is used to crop the object region from the original image. Then, the cropped image region is fed to the LMM to obtain the final answer. As shown in Figure~\ref{fig:sup_viscot}, the Visual CoT empowered by the LMM's grounding ability is helpful when the LMM needs to focus on question-related regions for visual perception and reasoning. We observe that the LMM can infer the location of text contents (\eg, fax number) in a document and the mask head can localise the relevant region, even though they are not trained on such data (\ie VisCoT~\cite{shao2024visualcot}).

\noindent\textbf{Attention Maps of Different LMMs.} In Figure~\ref{fig:sup_attn_maps}, we compare the attention maps of DeepSeekVL-7B(indicated as `D') and LLaVA-1.5-7B(indicated as `L'). (m,n) means attention map at the n-th head of the m-th layer.
The two models are similar in showcasing shapes and locations of objects but differ in specific attention map patterns and textures. For example, LLaVA's attention tends to be more concentrated with higher amplitudes.

\subsection{Broader Impact}
\label{sec:impact}

This paper addresses an important challenge in large multimodal models---improving the specialised performance while preserving the model's general capabilities.
By decoupling the grounding and conversational abilities, building upon the frozen LMMs, the proposed approach allows LMMs to visually ground objects and maintain their broad language capability.
Our work is expected to have extensive benefits: 
(1) It enables the deployment of visually grounding LMMs in real-world applications that require both specialised multimodal capabilities and general language understanding, such as assistive tools and interactive robotics. 
(2) It paves the way for more flexible and adaptable multimodal AI systems that can be tailored to specific tasks or domains without compromising their core language capabilities.
(3) Preserving instruction-following ability and resistance to hallucinations can improve the safety and reliability of the systems, making them suitable for high-stakes applications.

However, similar to many LMM-based systems, there are also potential negative impacts that should be considered: (1) Potential Bias: The pre-trained off-the-shelf LMMs used in the F-LMM approach may already contain biases, which could be propagated through the grounding process.
(2) Potential for Displacement of Human Labor: The increased capabilities of visually grounding LMMs could lead to the displacement of human labor in certain domains, such as customer service, content creation, or image analysis.
(3) Privacy and Ethical Concerns: Integrating visual grounding capabilities with language models raises privacy concerns, as the models could potentially be used to identify individuals or extract sensitive information from images.

To avoid misuse of the model, we will adopt the following safeguards:
1) Access Controls: Strict authentication and authorisation mechanisms will be implemented to ensure that only authorised and responsible individuals or organisations can access and use the models.
2) Usage Policies and Agreements: Clear usage policies and agreements will be established to define the intended purpose of the models. These policies will explicitly prohibit any malicious or harmful activities. Users will be required to agree to these policies and may face legal consequences if they violate them.
3) Transparency: We are committed to promoting transparency by providing comprehensive descriptions of the model's capabilities, limitations, the training pipeline, and the datasets used.

\subsection{Limitations}
\label{sec:limitation}
While the proposed F-LMM approach demonstrates promising results in preserving conversational abilities while enhancing visual grounding, there are several key limitations that warrant consideration.

\begin{itemize}
    \item Inherited Biases and Limitations: As the F-LMM method is built upon frozen pre-trained LMMs, it inherits any biases or limitations present in the underlying models. These could include demographic biases, skewed knowledge representations, or other undesirable properties.

\item  Limited Modality Scope: This work primarily focuses on vision-language multimodal interactions, without exploring other important modalities such as video, audio, and 3D point clouds. Expanding the scope to these additional modalities is a great direction to explore in the future.

\item Model Size Constraints: The experiments were restricted to LMMs up to 8 billion in parameter counts due to limited computing resources. Larger and more powerful models beyond this scale were not included. 
To address these limitations, future research could focus on mitigating biases, expanding the modality scope, and exploring larger-scale models.

\end{itemize}


\newpage
\noindent\textbf{Acknowledgement.} This research is supported by the National Research Foundation, Singapore under its AI Singapore Programme (AISG Award No: AISG3-PhD-2023-08-048T), the RIE2020 Industry Alignment Fund – Industry Collaboration Projects (IAF-ICP) Funding Initiative, as well as cash and in-kind contribution from the industry partner(s).

{
    \small
    \bibliographystyle{ieeenat_fullname}
    \bibliography{main}

\begin{thebibliography}{89}
\providecommand{\natexlab}[1]{#1}
\providecommand{\url}[1]{\texttt{#1}}
\expandafter\ifx\csname urlstyle\endcsname\relax
  \providecommand{\doi}[1]{doi: #1}\else
  \providecommand{\doi}{doi: \begingroup \urlstyle{rm}\Url}\fi

\bibitem[Achiam et~al.(2023)Achiam, Adler, Agarwal, Ahmad, Akkaya, Aleman, Almeida, Altenschmidt, Altman, Anadkat, et~al.]{achiam2023gpt4}
Josh Achiam, Steven Adler, Sandhini Agarwal, Lama Ahmad, Ilge Akkaya, Florencia~Leoni Aleman, Diogo Almeida, Janko Altenschmidt, Sam Altman, Shyamal Anadkat, et~al.
\newblock Gpt-4 technical report.
\newblock \emph{arXiv preprint arXiv:2303.08774}, 2023.

\bibitem[Alayrac et~al.(2022)Alayrac, Donahue, Luc, Miech, Barr, Hasson, Lenc, Mensch, Millican, Reynolds, et~al.]{alayrac2022flamingo}
Jean-Baptiste Alayrac, Jeff Donahue, Pauline Luc, Antoine Miech, Iain Barr, Yana Hasson, Karel Lenc, Arthur Mensch, Katherine Millican, Malcolm Reynolds, et~al.
\newblock Flamingo: a visual language model for few-shot learning.
\newblock \emph{Advances in neural information processing systems}, 35:\penalty0 23716--23736, 2022.

\bibitem[Bai et~al.(2023)Bai, Bai, Yang, Wang, Tan, Wang, Lin, Zhou, and Zhou]{bai2023qwenvl}
Jinze Bai, Shuai Bai, Shusheng Yang, Shijie Wang, Sinan Tan, Peng Wang, Junyang Lin, Chang Zhou, and Jingren Zhou.
\newblock Qwen-vl: A frontier large vision-language model with versatile abilities.
\newblock \emph{arXiv preprint arXiv:2308.12966}, 2023.

\bibitem[Brown et~al.(2020)Brown, Mann, Ryder, Subbiah, Kaplan, Dhariwal, Neelakantan, Shyam, Sastry, Askell, et~al.]{brown2020gpt3}
Tom Brown, Benjamin Mann, Nick Ryder, Melanie Subbiah, Jared~D Kaplan, Prafulla Dhariwal, Arvind Neelakantan, Pranav Shyam, Girish Sastry, Amanda Askell, et~al.
\newblock Language models are few-shot learners.
\newblock \emph{Advances in neural information processing systems}, 33:\penalty0 1877--1901, 2020.

\bibitem[Caesar et~al.(2018)Caesar, Uijlings, and Ferrari]{cocostuff}
Holger Caesar, Jasper Uijlings, and Vittorio Ferrari.
\newblock Coco-stuff: Thing and stuff classes in context.
\newblock In \emph{CVPR}, pages 1209--1218, 2018.

\bibitem[Chefer et~al.(2021)Chefer, Gur, and Wolf]{chefer2021generic}
Hila Chefer, Shir Gur, and Lior Wolf.
\newblock Generic attention-model explainability for interpreting bi-modal and encoder-decoder transformers.
\newblock In \emph{Proceedings of the IEEE/CVF International Conference on Computer Vision}, pages 397--406, 2021.

\bibitem[Chen et~al.(2023)Chen, Zhang, Zeng, Zhang, Zhu, and Zhao]{chen2023shikra}
Keqin Chen, Zhao Zhang, Weili Zeng, Richong Zhang, Feng Zhu, and Rui Zhao.
\newblock Shikra: Unleashing multimodal llm's referential dialogue magic.
\newblock \emph{arXiv preprint arXiv:2306.15195}, 2023.

\bibitem[Chen et~al.(2017)Chen, Papandreou, Kokkinos, Murphy, and Yuille]{chen2017deeplab}
Liang-Chieh Chen, George Papandreou, Iasonas Kokkinos, Kevin Murphy, and Alan~L Yuille.
\newblock Deeplab: Semantic image segmentation with deep convolutional nets, atrous convolution, and fully connected crfs.
\newblock \emph{IEEE transactions on pattern analysis and machine intelligence}, 40\penalty0 (4):\penalty0 834--848, 2017.

\bibitem[Cheng et~al.(2020)Cheng, Collins, Zhu, Liu, Huang, Adam, and Chen]{cheng2020panoptic_deeplab}
Bowen Cheng, Maxwell~D Collins, Yukun Zhu, Ting Liu, Thomas~S Huang, Hartwig Adam, and Liang-Chieh Chen.
\newblock Panoptic-deeplab: A simple, strong, and fast baseline for bottom-up panoptic segmentation.
\newblock In \emph{Proceedings of the IEEE/CVF conference on computer vision and pattern recognition}, pages 12475--12485, 2020.

\bibitem[Cheng et~al.(2021)Cheng, Schwing, and Kirillov]{cheng2021maskformer}
Bowen Cheng, Alex Schwing, and Alexander Kirillov.
\newblock Per-pixel classification is not all you need for semantic segmentation.
\newblock \emph{Advances in neural information processing systems}, 34:\penalty0 17864--17875, 2021.

\bibitem[Cheng et~al.(2022)Cheng, Misra, Schwing, Kirillov, and Girdhar]{cheng2022mask2former}
Bowen Cheng, Ishan Misra, Alexander~G Schwing, Alexander Kirillov, and Rohit Girdhar.
\newblock Masked-attention mask transformer for universal image segmentation.
\newblock In \emph{Proceedings of the IEEE/CVF conference on computer vision and pattern recognition}, pages 1290--1299, 2022.

\bibitem[Chowdhery et~al.(2023)Chowdhery, Narang, Devlin, Bosma, Mishra, Roberts, Barham, Chung, Sutton, Gehrmann, et~al.]{chowdhery2023palm}
Aakanksha Chowdhery, Sharan Narang, Jacob Devlin, Maarten Bosma, Gaurav Mishra, Adam Roberts, Paul Barham, Hyung~Won Chung, Charles Sutton, Sebastian Gehrmann, et~al.
\newblock Palm: Scaling language modeling with pathways.
\newblock \emph{Journal of Machine Learning Research}, 24\penalty0 (240):\penalty0 1--113, 2023.

\bibitem[Dai et~al.(2024)Dai, Li, Li, Tiong, Zhao, Wang, Li, Fung, and Hoi]{dai2024instructblip}
Wenliang Dai, Junnan Li, Dongxu Li, Anthony Meng~Huat Tiong, Junqi Zhao, Weisheng Wang, Boyang Li, Pascale~N Fung, and Steven Hoi.
\newblock Instructblip: Towards general-purpose vision-language models with instruction tuning.
\newblock \emph{Advances in Neural Information Processing Systems}, 36, 2024.

\bibitem[Ding et~al.(2022)Ding, Ding, Hui, Huang, Wei, Wei, and Liu]{ding2022ppmn}
Zihan Ding, Zi-han Ding, Tianrui Hui, Junshi Huang, Xiaoming Wei, Xiaolin Wei, and Si Liu.
\newblock Ppmn: Pixel-phrase matching network for one-stage panoptic narrative grounding.
\newblock In \emph{Proceedings of the 30th ACM International Conference on Multimedia}, pages 5537--5546, 2022.

\bibitem[Fu et~al.(2023)Fu, Chen, Shen, Qin, Zhang, Lin, Yang, Zheng, Li, Sun, Wu, and Ji]{fu2023mme}
Chaoyou Fu, Peixian Chen, Yunhang Shen, Yulei Qin, Mengdan Zhang, Xu Lin, Jinrui Yang, Xiawu Zheng, Ke Li, Xing Sun, Yunsheng Wu, and Rongrong Ji.
\newblock Mme: A comprehensive evaluation benchmark for multimodal large language models.
\newblock \emph{arXiv preprint arXiv:2306.13394}, 2023.

\bibitem[Gonz{\'a}lez et~al.(2021)Gonz{\'a}lez, Ayobi, Hern{\'a}ndez, Hern{\'a}ndez, Pont-Tuset, and Arbel{\'a}ez]{png}
Cristina Gonz{\'a}lez, Nicol{\'a}s Ayobi, Isabela Hern{\'a}ndez, Jos{\'e} Hern{\'a}ndez, Jordi Pont-Tuset, and Pablo Arbel{\'a}ez.
\newblock Panoptic narrative grounding.
\newblock In \emph{Proceedings of the IEEE/CVF International Conference on Computer Vision}, pages 1364--1373, 2021.

\bibitem[Guo et~al.(2024)Guo, Wang, Ma, Ji, and Sun]{guo2024xpng}
Tianyu Guo, Haowei Wang, Yiwei Ma, Jiayi Ji, and Xiaoshuai Sun.
\newblock Improving panoptic narrative grounding by harnessing semantic relationships and visual confirmation.
\newblock In \emph{Proceedings of the AAAI Conference on Artificial Intelligence}, pages 1985--1993, 2024.

\bibitem[He et~al.(2022)He, Yang, Yang, Kortylewski, Yuan, Chen, Liu, Yang, Yu, and Yuille]{he2022partimagenet}
Ju He, Shuo Yang, Shaokang Yang, Adam Kortylewski, Xiaoding Yuan, Jie-Neng Chen, Shuai Liu, Cheng Yang, Qihang Yu, and Alan Yuille.
\newblock Partimagenet: A large, high-quality dataset of parts.
\newblock In \emph{European Conference on Computer Vision}, pages 128--145. Springer, 2022.

\bibitem[He et~al.(2017)He, Gkioxari, Doll{\'a}r, and Girshick]{he2017mask}
Kaiming He, Georgia Gkioxari, Piotr Doll{\'a}r, and Ross Girshick.
\newblock Mask r-cnn.
\newblock In \emph{Proceedings of the IEEE international conference on computer vision}, pages 2961--2969, 2017.

\bibitem[Honnibal and Montani(2017)]{spacy2}
Matthew Honnibal and Ines Montani.
\newblock {spaCy 2}: Natural language understanding with {B}loom embeddings, convolutional neural networks and incremental parsing.
\newblock To appear, 2017.

\bibitem[Huang et~al.(2024)Huang, Dong, Zhang, Wang, He, Wang, Lin, Zhang, and Yu]{huang2024opera}
Qidong Huang, Xiaoyi Dong, Pan Zhang, Bin Wang, Conghui He, Jiaqi Wang, Dahua Lin, Weiming Zhang, and Nenghai Yu.
\newblock Opera: Alleviating hallucination in multi-modal large language models via over-trust penalty and retrospection-allocation.
\newblock In \emph{Proceedings of the IEEE/CVF Conference on Computer Vision and Pattern Recognition}, pages 13418--13427, 2024.

\bibitem[Jiang et~al.(2023)Jiang, Sablayrolles, Mensch, Bamford, Chaplot, Casas, Bressand, Lengyel, Lample, Saulnier, et~al.]{jiang2023mistral}
Albert~Q Jiang, Alexandre Sablayrolles, Arthur Mensch, Chris Bamford, Devendra~Singh Chaplot, Diego de~las Casas, Florian Bressand, Gianna Lengyel, Guillaume Lample, Lucile Saulnier, et~al.
\newblock Mistral 7b.
\newblock \emph{arXiv preprint arXiv:2310.06825}, 2023.

\bibitem[Kazemzadeh et~al.(2014{\natexlab{a}})Kazemzadeh, Ordonez, Matten, and Berg]{kazemzadeh2014referitgame}
Sahar Kazemzadeh, Vicente Ordonez, Mark Matten, and Tamara Berg.
\newblock Referitgame: Referring to objects in photographs of natural scenes.
\newblock In \emph{Proceedings of the 2014 conference on empirical methods in natural language processing (EMNLP)}, pages 787--798, 2014{\natexlab{a}}.

\bibitem[Kazemzadeh et~al.(2014{\natexlab{b}})Kazemzadeh, Ordonez, Matten, and Berg]{refcoco}
Sahar Kazemzadeh, Vicente Ordonez, Mark Matten, and Tamara Berg.
\newblock Referitgame: Referring to objects in photographs of natural scenes.
\newblock In \emph{Proceedings of the 2014 conference on empirical methods in natural language processing (EMNLP)}, pages 787--798, 2014{\natexlab{b}}.

\bibitem[Kirillov et~al.(2019)Kirillov, He, Girshick, Rother, and Doll{\'{a}}r]{panoptic}
Alexander Kirillov, Kaiming He, Ross~B. Girshick, Carsten Rother, and Piotr Doll{\'{a}}r.
\newblock Panoptic segmentation.
\newblock \emph{CVPR}, 2019.

\bibitem[Kirillov et~al.(2023)Kirillov, Mintun, Ravi, Mao, Rolland, Gustafson, Xiao, Whitehead, Berg, Lo, et~al.]{kirillov2023sam}
Alexander Kirillov, Eric Mintun, Nikhila Ravi, Hanzi Mao, Chloe Rolland, Laura Gustafson, Tete Xiao, Spencer Whitehead, Alexander~C Berg, Wan-Yen Lo, et~al.
\newblock Segment anything.
\newblock \emph{arXiv preprint arXiv:2304.02643}, 2023.

\bibitem[Krishna et~al.(2017)Krishna, Zhu, Groth, Johnson, Hata, Kravitz, Chen, Kalantidis, Li, Shamma, Bernstein, and Fei-Fei]{visualgenome}
Ranjay Krishna, Yuke Zhu, Oliver Groth, Justin Johnson, Kenji Hata, Joshua Kravitz, Stephanie Chen, Yannis Kalantidis, Li-Jia Li, David~A. Shamma, Michael~S. Bernstein, and Li Fei-Fei.
\newblock {Visual Genome: Connecting} language and vision using crowdsourced dense image annotations.
\newblock \emph{IJCV}, 2017.

\bibitem[Lai et~al.(2023)Lai, Tian, Chen, Li, Yuan, Liu, and Jia]{lai2023lisa}
Xin Lai, Zhuotao Tian, Yukang Chen, Yanwei Li, Yuhui Yuan, Shu Liu, and Jiaya Jia.
\newblock Lisa: Reasoning segmentation via large language model.
\newblock \emph{arXiv preprint arXiv:2308.00692}, 2023.

\bibitem[Lauren{\c{c}}on et~al.(2024)Lauren{\c{c}}on, Tronchon, Cord, and Sanh]{laurenccon2024matters}
Hugo Lauren{\c{c}}on, L{\'e}o Tronchon, Matthieu Cord, and Victor Sanh.
\newblock What matters when building vision-language models?
\newblock \emph{arXiv preprint arXiv:2405.02246}, 2024.

\bibitem[Li et~al.(2023{\natexlab{a}})Li, Zhang, Chen, Wang, Pu, Yang, Li, and Liu]{li2023otter}
Bo Li, Yuanhan Zhang, Liangyu Chen, Jinghao Wang, Fanyi Pu, Jingkang Yang, Chunyuan Li, and Ziwei Liu.
\newblock Mimic-it: Multi-modal in-context instruction tuning.
\newblock \emph{arXiv preprint arXiv:2306.05425}, 2023{\natexlab{a}}.

\bibitem[Li et~al.(2023{\natexlab{b}})Li, Li, Savarese, and Hoi]{li2023blip2}
Junnan Li, Dongxu Li, Silvio Savarese, and Steven Hoi.
\newblock Blip-2: Bootstrapping language-image pre-training with frozen image encoders and large language models.
\newblock In \emph{International conference on machine learning}, pages 19730--19742. PMLR, 2023{\natexlab{b}}.

\bibitem[Li et~al.(2021)Li, Zhao, Qi, Wang, Li, Sun, and Jia]{li2021fcn4panoptic_seg}
Yanwei Li, Hengshuang Zhao, Xiaojuan Qi, Liwei Wang, Zeming Li, Jian Sun, and Jiaya Jia.
\newblock Fully convolutional networks for panoptic segmentation.
\newblock In \emph{Proceedings of the IEEE/CVF conference on computer vision and pattern recognition}, pages 214--223, 2021.

\bibitem[Li et~al.(2023{\natexlab{c}})Li, Du, Zhou, Wang, Zhao, and Wen]{li2023pope}
Yifan Li, Yifan Du, Kun Zhou, Jinpeng Wang, Wayne~Xin Zhao, and Ji-Rong Wen.
\newblock Evaluating object hallucination in large vision-language models.
\newblock \emph{arXiv preprint arXiv:2305.10355}, 2023{\natexlab{c}}.

\bibitem[Li et~al.(2023{\natexlab{d}})Li, Zhang, Wang, Zhong, Chen, Chu, Liu, and Jia]{li2024mgm}
Yanwei Li, Yuechen Zhang, Chengyao Wang, Zhisheng Zhong, Yixin Chen, Ruihang Chu, Shaoteng Liu, and Jiaya Jia.
\newblock Mini-gemini: Mining the potential of multi-modality vision language models.
\newblock \emph{arXiv:2403.18814}, 2023{\natexlab{d}}.

\bibitem[Liang et~al.(2023)Liang, Wu, Dai, Li, Zhao, Zhang, Zhang, Vajda, and Marculescu]{liang2023open}
Feng Liang, Bichen Wu, Xiaoliang Dai, Kunpeng Li, Yinan Zhao, Hang Zhang, Peizhao Zhang, Peter Vajda, and Diana Marculescu.
\newblock Open-vocabulary semantic segmentation with mask-adapted clip.
\newblock In \emph{Proceedings of the IEEE/CVF Conference on Computer Vision and Pattern Recognition}, pages 7061--7070, 2023.

\bibitem[Lin et~al.(2023)Lin, Yin, Ping, Lu, Molchanov, Tao, Mao, Kautz, Shoeybi, and Han]{lin2023vila}
Ji Lin, Hongxu Yin, Wei Ping, Yao Lu, Pavlo Molchanov, Andrew Tao, Huizi Mao, Jan Kautz, Mohammad Shoeybi, and Song Han.
\newblock Vila: On pre-training for visual language models.
\newblock \emph{arXiv preprint arXiv:2312.07533}, 2023.

\bibitem[Liu et~al.(2023{\natexlab{a}})Liu, Ding, and Jiang]{liu2023gres}
Chang Liu, Henghui Ding, and Xudong Jiang.
\newblock Gres: Generalized referring expression segmentation.
\newblock In \emph{Proceedings of the IEEE/CVF conference on computer vision and pattern recognition}, pages 23592--23601, 2023{\natexlab{a}}.

\bibitem[Liu et~al.(2023{\natexlab{b}})Liu, Li, Li, and Lee]{liu2023improvedllava}
Haotian Liu, Chunyuan Li, Yuheng Li, and Yong~Jae Lee.
\newblock Improved baselines with visual instruction tuning, 2023{\natexlab{b}}.

\bibitem[Liu et~al.(2023{\natexlab{c}})Liu, Li, Wu, and Lee]{liu2023llava}
Haotian Liu, Chunyuan Li, Qingyang Wu, and Yong~Jae Lee.
\newblock Visual instruction tuning, 2023{\natexlab{c}}.

\bibitem[Liu et~al.(2024{\natexlab{a}})Liu, Li, Li, Li, Zhang, Shen, and Lee]{liu2024llavanext}
Haotian Liu, Chunyuan Li, Yuheng Li, Bo Li, Yuanhan Zhang, Sheng Shen, and Yong~Jae Lee.
\newblock Llava-next: Improved reasoning, ocr, and world knowledge, 2024{\natexlab{a}}.

\bibitem[Liu et~al.(2023{\natexlab{d}})Liu, Zeng, Ren, Li, Zhang, Yang, Li, Yang, Su, Zhu, et~al.]{liu2023grounding}
Shilong Liu, Zhaoyang Zeng, Tianhe Ren, Feng Li, Hao Zhang, Jie Yang, Chunyuan Li, Jianwei Yang, Hang Su, Jun Zhu, et~al.
\newblock Grounding dino: Marrying dino with grounded pre-training for open-set object detection.
\newblock \emph{arXiv preprint arXiv:2303.05499}, 2023{\natexlab{d}}.

\bibitem[Liu et~al.(2024{\natexlab{b}})Liu, Duan, Zhang, Li, Zhang, Zhao, Yuan, Wang, He, Liu, Chen, and Lin]{liu2024mmbench}
Yuan Liu, Haodong Duan, Yuanhan Zhang, Bo Li, Songyang Zhang, Wangbo Zhao, Yike Yuan, Jiaqi Wang, Conghui He, Ziwei Liu, Kai Chen, and Dahua Lin.
\newblock Mmbench: Is your multi-modal model an all-around player?, 2024{\natexlab{b}}.

\bibitem[Loshchilov and Hutter(2019)]{loshchilov2019decoupled}
Ilya Loshchilov and Frank Hutter.
\newblock Decoupled weight decay regularization, 2019.

\bibitem[Lu et~al.(2024)Lu, Liu, Zhang, Wang, Dong, Liu, Sun, Ren, Li, Yang, Sun, Deng, Xu, Xie, and Ruan]{lu2024deepseekvl}
Haoyu Lu, Wen Liu, Bo Zhang, Bingxuan Wang, Kai Dong, Bo Liu, Jingxiang Sun, Tongzheng Ren, Zhuoshu Li, Hao Yang, Yaofeng Sun, Chengqi Deng, Hanwei Xu, Zhenda Xie, and Chong Ruan.
\newblock Deepseek-vl: Towards real-world vision-language understanding, 2024.

\bibitem[Luo et~al.(2020)Luo, Zhou, Sun, Cao, Wu, Deng, and Ji]{luo2020mcn}
Gen Luo, Yiyi Zhou, Xiaoshuai Sun, Liujuan Cao, Chenglin Wu, Cheng Deng, and Rongrong Ji.
\newblock Multi-task collaborative network for joint referring expression comprehension and segmentation.
\newblock In \emph{Proceedings of the IEEE/CVF Conference on computer vision and pattern recognition}, pages 10034--10043, 2020.

\bibitem[Mao et~al.(2016)Mao, Huang, Toshev, Camburu, Yuille, and Murphy]{refcocog}
Junhua Mao, Jonathan Huang, Alexander Toshev, Oana Camburu, Alan~L Yuille, and Kevin Murphy.
\newblock Generation and comprehension of unambiguous object descriptions.
\newblock In \emph{Proceedings of the IEEE conference on computer vision and pattern recognition}, pages 11--20, 2016.

\bibitem[McKinzie et~al.(2024)McKinzie, Gan, Fauconnier, Dodge, Zhang, Dufter, Shah, Du, Peng, Weers, et~al.]{mckinzie2024mm1}
Brandon McKinzie, Zhe Gan, Jean-Philippe Fauconnier, Sam Dodge, Bowen Zhang, Philipp Dufter, Dhruti Shah, Xianzhi Du, Futang Peng, Floris Weers, et~al.
\newblock Mm1: Methods, analysis \& insights from multimodal llm pre-training.
\newblock \emph{arXiv preprint arXiv:2403.09611}, 2024.

\bibitem[Mesnard et~al.(2024)Mesnard, Hardin, Dadashi, Bhupatiraju, Pathak, Sifre, Rivi{\`e}re, Kale, Love, et~al.]{mesnard2024gemma}
Thomas Mesnard, Cassidy Hardin, Robert Dadashi, Surya Bhupatiraju, Shreya Pathak, Laurent Sifre, Morgane Rivi{\`e}re, Mihir~Sanjay Kale, Juliette Love, et~al.
\newblock Gemma: Open models based on gemini research and technology.
\newblock \emph{arXiv preprint arXiv:2403.08295}, 2024.

\bibitem[Nagaraja et~al.(2016)Nagaraja, Morariu, and Davis]{nagaraja2016modeling_ref}
Varun~K Nagaraja, Vlad~I Morariu, and Larry~S Davis.
\newblock Modeling context between objects for referring expression understanding.
\newblock In \emph{Computer Vision--ECCV 2016: 14th European Conference, Amsterdam, The Netherlands, October 11--14, 2016, Proceedings, Part IV 14}, pages 792--807. Springer, 2016.

\bibitem[Ouyang et~al.(2022)Ouyang, Wu, Jiang, Almeida, Wainwright, Mishkin, Zhang, Agarwal, Slama, Ray, et~al.]{ouyang2022hf}
Long Ouyang, Jeffrey Wu, Xu Jiang, Diogo Almeida, Carroll Wainwright, Pamela Mishkin, Chong Zhang, Sandhini Agarwal, Katarina Slama, Alex Ray, et~al.
\newblock Training language models to follow instructions with human feedback.
\newblock \emph{Advances in neural information processing systems}, 35:\penalty0 27730--27744, 2022.

\bibitem[Peng et~al.(2023{\natexlab{a}})Peng, Li, He, Galley, and Gao]{vicuna}
Baolin Peng, Chunyuan Li, Pengcheng He, Michel Galley, and Jianfeng Gao.
\newblock Instruction tuning with gpt-4.
\newblock \emph{arXiv preprint arXiv:2304.03277}, 2023{\natexlab{a}}.

\bibitem[Peng et~al.(2023{\natexlab{b}})Peng, Wang, Dong, Hao, Huang, Ma, and Wei]{peng2023kosmos}
Zhiliang Peng, Wenhui Wang, Li Dong, Yaru Hao, Shaohan Huang, Shuming Ma, and Furu Wei.
\newblock Kosmos-2: Grounding multimodal large language models to the world.
\newblock \emph{arXiv preprint arXiv:2306.14824}, 2023{\natexlab{b}}.

\bibitem[Peng et~al.(2023{\natexlab{c}})Peng, Wang, Dong, Hao, Huang, Ma, and Wei]{peng2023kosmos2}
Zhiliang Peng, Wenhui Wang, Li Dong, Yaru Hao, Shaohan Huang, Shuming Ma, and Furu Wei.
\newblock Kosmos-2: Grounding multimodal large language models to the world.
\newblock \emph{arXiv preprint arXiv:2306.14824}, 2023{\natexlab{c}}.

\bibitem[Pi et~al.(2023)Pi, Yao, Gao, Zhang, and Zhang]{pi2023perceptiongpt}
Renjie Pi, Lewei Yao, Jiahui Gao, Jipeng Zhang, and Tong Zhang.
\newblock Perceptiongpt: Effectively fusing visual perception into llm.
\newblock \emph{arXiv preprint arXiv:2311.06612}, 2023.

\bibitem[Plummer et~al.(2015)Plummer, Wang, Cervantes, Caicedo, Hockenmaier, and Lazebnik]{plummer2015flickr30k}
Bryan~A Plummer, Liwei Wang, Chris~M Cervantes, Juan~C Caicedo, Julia Hockenmaier, and Svetlana Lazebnik.
\newblock Flickr30k entities: Collecting region-to-phrase correspondences for richer image-to-sentence models.
\newblock In \emph{ICCV}, pages 2641--2649, 2015.

\bibitem[Radford et~al.(2018)Radford, Narasimhan, Salimans, Sutskever, et~al.]{radford2018gpt}
Alec Radford, Karthik Narasimhan, Tim Salimans, Ilya Sutskever, et~al.
\newblock Improving language understanding by generative pre-training.
\newblock 2018.

\bibitem[Radford et~al.(2019)Radford, Wu, Child, Luan, Amodei, Sutskever, et~al.]{radford2019gpt2}
Alec Radford, Jeffrey Wu, Rewon Child, David Luan, Dario Amodei, Ilya Sutskever, et~al.
\newblock Language models are unsupervised multitask learners.
\newblock \emph{OpenAI blog}, 1\penalty0 (8):\penalty0 9, 2019.

\bibitem[Radford et~al.(2021)Radford, Kim, Hallacy, Ramesh, Goh, Agarwal, Sastry, Askell, Mishkin, Clark, et~al.]{radford2021learning}
Alec Radford, Jong~Wook Kim, Chris Hallacy, Aditya Ramesh, Gabriel Goh, Sandhini Agarwal, Girish Sastry, Amanda Askell, Pamela Mishkin, Jack Clark, et~al.
\newblock Learning transferable visual models from natural language supervision.
\newblock 2021.

\bibitem[Ramanathan et~al.(2023)Ramanathan, Kalia, Petrovic, Wen, Zheng, Guo, Wang, Marquez, Kovvuri, Kadian, et~al.]{ramanathan2023paco}
Vignesh Ramanathan, Anmol Kalia, Vladan Petrovic, Yi Wen, Baixue Zheng, Baishan Guo, Rui Wang, Aaron Marquez, Rama Kovvuri, Abhishek Kadian, et~al.
\newblock Paco: Parts and attributes of common objects.
\newblock In \emph{Proceedings of the IEEE/CVF Conference on Computer Vision and Pattern Recognition}, pages 7141--7151, 2023.

\bibitem[Rasheed et~al.(2024)Rasheed, Maaz, Shaji, Shaker, Khan, Cholakkal, Anwer, Xing, Yang, and Khan]{hanoona2023GLaMM}
Hanoona Rasheed, Muhammad Maaz, Sahal Shaji, Abdelrahman Shaker, Salman Khan, Hisham Cholakkal, Rao~M. Anwer, Eric Xing, Ming-Hsuan Yang, and Fahad~S. Khan.
\newblock Glamm: Pixel grounding large multimodal model.
\newblock \emph{The IEEE/CVF Conference on Computer Vision and Pattern Recognition}, 2024.

\bibitem[Ren et~al.(2023)Ren, Huang, Wei, Zhao, Fu, Feng, and Jin]{ren2023pixellm}
Zhongwei Ren, Zhicheng Huang, Yunchao Wei, Yao Zhao, Dongmei Fu, Jiashi Feng, and Xiaojie Jin.
\newblock Pixellm: Pixel reasoning with large multimodal model.
\newblock \emph{arXiv preprint arXiv:2312.02228}, 2023.

\bibitem[Ronneberger et~al.(2015)Ronneberger, Fischer, and Brox]{ronneberger2015u}
Olaf Ronneberger, Philipp Fischer, and Thomas Brox.
\newblock U-net: Convolutional networks for biomedical image segmentation.
\newblock In \emph{Medical image computing and computer-assisted intervention--MICCAI 2015: 18th international conference, Munich, Germany, October 5-9, 2015, proceedings, part III 18}, pages 234--241. Springer, 2015.

\bibitem[Shao et~al.(2024)Shao, Qian, Xiao, Song, Zong, Wang, Liu, and Li]{shao2024visualcot}
Hao Shao, Shengju Qian, Han Xiao, Guanglu Song, Zhuofan Zong, Letian Wang, Yu Liu, and Hongsheng Li.
\newblock Visual cot: Unleashing chain-of-thought reasoning in multi-modal language models.
\newblock \emph{arXiv preprint arXiv:2403.16999}, 2024.

\bibitem[Sudre et~al.(2017)Sudre, Li, Vercauteren, Ourselin, and Jorge~Cardoso]{sudre2017dice}
Carole~H Sudre, Wenqi Li, Tom Vercauteren, Sebastien Ourselin, and M Jorge~Cardoso.
\newblock Generalised dice overlap as a deep learning loss function for highly unbalanced segmentations.
\newblock In \emph{Deep Learning in Medical Image Analysis and Multimodal Learning for Clinical Decision Support: Third International Workshop, DLMIA 2017, and 7th International Workshop, ML-CDS 2017, Held in Conjunction with MICCAI 2017, Qu{\'e}bec City, QC, Canada, September 14, Proceedings 3}, pages 240--248. Springer, 2017.

\bibitem[Team(2024)]{hpt}
HyperGAI Team.
\newblock Hpt 1.5 air: Best open-sourced 8b multimodal llm with llama 3, 2024.

\bibitem[Touvron et~al.(2023{\natexlab{a}})Touvron, Lavril, Izacard, Martinet, Lachaux, Lacroix, Rozi{\`e}re, Goyal, Hambro, Azhar, et~al.]{llama}
Hugo Touvron, Thibaut Lavril, Gautier Izacard, Xavier Martinet, Marie-Anne Lachaux, Timoth{\'e}e Lacroix, Baptiste Rozi{\`e}re, Naman Goyal, Eric Hambro, Faisal Azhar, et~al.
\newblock Llama: Open and efficient foundation language models.
\newblock \emph{arXiv preprint arXiv:2302.13971}, 2023{\natexlab{a}}.

\bibitem[Touvron et~al.(2023{\natexlab{b}})Touvron, Martin, Stone, Albert, Almahairi, Babaei, Bashlykov, Batra, Bhargava, Bhosale, et~al.]{llama2}
Hugo Touvron, Louis Martin, Kevin Stone, Peter Albert, Amjad Almahairi, Yasmine Babaei, Nikolay Bashlykov, Soumya Batra, Prajjwal Bhargava, Shruti Bhosale, et~al.
\newblock Llama 2: Open foundation and fine-tuned chat models.
\newblock \emph{arXiv preprint arXiv:2307.09288}, 2023{\natexlab{b}}.

\bibitem[Vaswani et~al.(2017)Vaswani, Shazeer, Parmar, Uszkoreit, Jones, Gomez, Kaiser, and Polosukhin]{transformer}
Ashish Vaswani, Noam Shazeer, Niki Parmar, Jakob Uszkoreit, Llion Jones, Aidan~N. Gomez, Lukasz Kaiser, and Illia Polosukhin.
\newblock Attention is all you need.
\newblock In \emph{NeurIPS}, 2017.

\bibitem[Wang et~al.(2023)Wang, Ji, Zhou, Wu, and Sun]{wang2023epng}
Haowei Wang, Jiayi Ji, Yiyi Zhou, Yongjian Wu, and Xiaoshuai Sun.
\newblock Towards real-time panoptic narrative grounding by an end-to-end grounding network.
\newblock In \emph{Proceedings of the AAAI Conference on Artificial Intelligence}, pages 2528--2536, 2023.

\bibitem[Wei et~al.(2024)Wei, Tan, Zhong, Yang, and Ma]{wei2024lasagna}
Cong Wei, Haoxian Tan, Yujie Zhong, Yujiu Yang, and Lin Ma.
\newblock Lasagna: Language-based segmentation assistant for complex queries.
\newblock \emph{arXiv preprint arXiv:2404.08506}, 2024.

\bibitem[Wei et~al.(2021)Wei, Bosma, Zhao, Guu, Yu, Lester, Du, Dai, and Le]{wei2021finetuned}
Jason Wei, Maarten Bosma, Vincent~Y Zhao, Kelvin Guu, Adams~Wei Yu, Brian Lester, Nan Du, Andrew~M Dai, and Quoc~V Le.
\newblock Finetuned language models are zero-shot learners.
\newblock \emph{arXiv preprint arXiv:2109.01652}, 2021.

\bibitem[Xia et~al.(2024)Xia, Han, Han, Pan, Song, and Huang]{xia2024gsva}
Zhuofan Xia, Dongchen Han, Yizeng Han, Xuran Pan, Shiji Song, and Gao Huang.
\newblock Gsva: Generalized segmentation via multimodal large language models.
\newblock In \emph{Proceedings of the IEEE/CVF Conference on Computer Vision and Pattern Recognition}, pages 3858--3869, 2024.

\bibitem[Xiong et~al.(2019)Xiong, Liao, Zhao, Hu, Bai, Yumer, and Urtasun]{xiong2019upsnet_panopticseg}
Yuwen Xiong, Renjie Liao, Hengshuang Zhao, Rui Hu, Min Bai, Ersin Yumer, and Raquel Urtasun.
\newblock Upsnet: A unified panoptic segmentation network.
\newblock In \emph{Proceedings of the IEEE/CVF conference on computer vision and pattern recognition}, pages 8818--8826, 2019.

\bibitem[Yang et~al.(2022)Yang, Wang, Tang, Chen, Zhao, and Torr]{yang2022lavt}
Zhao Yang, Jiaqi Wang, Yansong Tang, Kai Chen, Hengshuang Zhao, and Philip~HS Torr.
\newblock Lavt: Language-aware vision transformer for referring image segmentation.
\newblock In \emph{Proceedings of the IEEE/CVF Conference on Computer Vision and Pattern Recognition}, pages 18155--18165, 2022.

\bibitem[Ye et~al.(2023)Ye, Xu, Xu, Ye, Yan, Zhou, Wang, Hu, Shi, Shi, et~al.]{ye2023mplug}
Qinghao Ye, Haiyang Xu, Guohai Xu, Jiabo Ye, Ming Yan, Yiyang Zhou, Junyang Wang, Anwen Hu, Pengcheng Shi, Yaya Shi, et~al.
\newblock mplug-owl: Modularization empowers large language models with multimodality.
\newblock \emph{arXiv preprint arXiv:2304.14178}, 2023.

\bibitem[You et~al.(2023)You, Zhang, Gan, Du, Zhang, Wang, Cao, Chang, and Yang]{you2023ferret}
Haoxuan You, Haotian Zhang, Zhe Gan, Xianzhi Du, Bowen Zhang, Zirui Wang, Liangliang Cao, Shih-Fu Chang, and Yinfei Yang.
\newblock Ferret: Refer and ground anything anywhere at any granularity.
\newblock \emph{arXiv preprint arXiv:2310.07704}, 2023.

\bibitem[Yu et~al.(2024)Yu, Yang, Li, Wang, Lin, Liu, Wang, and Wang]{yu2024mmvet}
Weihao Yu, Zhengyuan Yang, Linjie Li, Jianfeng Wang, Kevin Lin, Zicheng Liu, Xinchao Wang, and Lijuan Wang.
\newblock Mm-vet: Evaluating large multimodal models for integrated capabilities.
\newblock In \emph{International conference on machine learning}. PMLR, 2024.

\bibitem[Zang et~al.(2023)Zang, Li, Han, Zhou, and Loy]{zang2023contextual}
Yuhang Zang, Wei Li, Jun Han, Kaiyang Zhou, and Chen~Change Loy.
\newblock Contextual object detection with multimodal large language models.
\newblock \emph{arXiv preprint arXiv:2305.18279}, 2023.

\bibitem[Zhai et~al.(2023)Zhai, Mustafa, Kolesnikov, and Beyer]{zhai2023sigmoid}
Xiaohua Zhai, Basil Mustafa, Alexander Kolesnikov, and Lucas Beyer.
\newblock Sigmoid loss for language image pre-training.
\newblock In \emph{Proceedings of the IEEE/CVF International Conference on Computer Vision}, pages 11975--11986, 2023.

\bibitem[Zhang et~al.(2023)Zhang, Li, Li, Ren, Zou, Liu, Huang, Gao, Zhang, Li, and Yang]{zhang2023llavagrounding}
Hao Zhang, Hongyang Li, Feng Li, Tianhe Ren, Xueyan Zou, Shilong Liu, Shijia Huang, Jianfeng Gao, Lei Zhang, Chunyuan Li, and Jianwei Yang.
\newblock Llava-grounding: Grounded visual chat with large multimodal models, 2023.

\bibitem[Zhang et~al.(2022)Zhang, Roller, Goyal, Artetxe, Chen, Chen, Dewan, Diab, Li, Lin, et~al.]{zhang2022opt}
Susan Zhang, Stephen Roller, Naman Goyal, Mikel Artetxe, Moya Chen, Shuohui Chen, Christopher Dewan, Mona Diab, Xian Li, Xi~Victoria Lin, et~al.
\newblock Opt: Open pre-trained transformer language models.
\newblock \emph{arXiv preprint arXiv:2205.01068}, 2022.

\bibitem[Zhang et~al.(2024{\natexlab{a}})Zhang, Li, Fei, Yuan, Wu, Ji, Loy, and Yan]{zhang2024omg}
Tao Zhang, Xiangtai Li, Hao Fei, Haobo Yuan, Shengqiong Wu, Shunping Ji, Chen~Change Loy, and Shuicheng Yan.
\newblock Omg-llava: Bridging image-level, object-level, pixel-level reasoning and understanding.
\newblock \emph{arXiv preprint arXiv:2406.19389}, 2024{\natexlab{a}}.

\bibitem[Zhang et~al.(2021)Zhang, Pang, Chen, and Loy]{zhang2021knet}
Wenwei Zhang, Jiangmiao Pang, Kai Chen, and Chen~Change Loy.
\newblock K-net: Towards unified image segmentation.
\newblock \emph{Advances in Neural Information Processing Systems}, 34:\penalty0 10326--10338, 2021.

\bibitem[Zhang et~al.(2024{\natexlab{b}})Zhang, Ma, Gao, Shakiah, Gao, and Chai]{zhang2024groundhog}
Yichi Zhang, Ziqiao Ma, Xiaofeng Gao, Suhaila Shakiah, Qiaozi Gao, and Joyce Chai.
\newblock Groundhog: Grounding large language models to holistic segmentation.
\newblock \emph{arXiv preprint arXiv:2402.16846}, 2024{\natexlab{b}}.

\bibitem[Zhang et~al.(2024{\natexlab{c}})Zhang, Ma, Zhang, and Bai]{zhang2024psalm}
Zheng Zhang, Yeyao Ma, Enming Zhang, and Xiang Bai.
\newblock Psalm: Pixelwise segmentation with large multi-modal model.
\newblock In \emph{European Conference on Computer Vision}, pages 74--91. Springer, 2024{\natexlab{c}}.

\bibitem[Zhao et~al.(2023)Zhao, Lin, Zhou, Huang, Feng, and Kang]{zhao2023bubogpt}
Yang Zhao, Zhijie Lin, Daquan Zhou, Zilong Huang, Jiashi Feng, and Bingyi Kang.
\newblock Bubogpt: Enabling visual grounding in multi-modal llms.
\newblock \emph{arXiv preprint arXiv:2307.08581}, 2023.

\bibitem[Zhou et~al.(2017)Zhou, Zhao, Puig, Fidler, Barriuso, and Torralba]{zhou2017ade20k}
Bolei Zhou, Hang Zhao, Xavier Puig, Sanja Fidler, Adela Barriuso, and Antonio Torralba.
\newblock Scene parsing through ade20k dataset.
\newblock In \emph{Proceedings of the IEEE conference on computer vision and pattern recognition}, pages 633--641, 2017.

\bibitem[Zou et~al.(2023)Zou, Dou, Yang, Gan, Li, Li, Dai, Behl, Wang, Yuan, et~al.]{zou2023generalized}
Xueyan Zou, Zi-Yi Dou, Jianwei Yang, Zhe Gan, Linjie Li, Chunyuan Li, Xiyang Dai, Harkirat Behl, Jianfeng Wang, Lu Yuan, et~al.
\newblock Generalized decoding for pixel, image, and language.
\newblock In \emph{Proceedings of the IEEE/CVF Conference on Computer Vision and Pattern Recognition}, pages 15116--15127, 2023.

\bibitem[Zou et~al.(2024)Zou, Yang, Zhang, Li, Li, Wang, Wang, Gao, and Lee]{zou2024seem}
Xueyan Zou, Jianwei Yang, Hao Zhang, Feng Li, Linjie Li, Jianfeng Wang, Lijuan Wang, Jianfeng Gao, and Yong~Jae Lee.
\newblock Segment everything everywhere all at once.
\newblock \emph{Advances in Neural Information Processing Systems}, 36, 2024.

\end{thebibliography}
}


\end{document}